\documentclass[journal]{IEEEtran}

\usepackage{times}
\usepackage{epsfig}
\usepackage{graphicx}
\usepackage{amsmath}
\usepackage{amssymb}
\usepackage{subfigure}
\usepackage{mathtools}
\usepackage{mathrsfs}
\usepackage{multirow}
\usepackage{stfloats}
\usepackage{psfrag}
\usepackage{algorithm}
\usepackage{algpseudocode}
\usepackage{booktabs}
\usepackage{color}
\usepackage[colorlinks,
            linkcolor=blue,       
            anchorcolor=blue,  
            citecolor=blue,       
            ]{hyperref}

\ifCLASSINFOpdf
\else
\fi
\begin{document}
%
\title{Vision-Language Matching for Text-to-Image Synthesis via Generative Adversarial Networks }
%
%
%

\author{{Qingrong Cheng,
Keyu Wen, and
Xiaodong Gu 
}
\thanks{Qingrong Cheng,
Keyu Wen, and
Xiaodong Gu
are with Department of Electronic Engineering,
Fudan University, Shanghai 200433, China (corresponding author: Xiaodong Gu;email: xdgu@fudan.edu.cn).
This work was supported by National Natural Science Foundation of China under grants 61771145 and 61371148.}
}

\maketitle

\begin{abstract}
Text-to-image synthesis is an attractive but challenging task that aims to generate a photo-realistic and semantic consistent image from a specific text description. The images synthesized by off-the-shelf models usually contain limited components compared with the corresponding image and text description, which decreases the image quality and the textual-visual consistency. To address this issue, we propose a novel Vision-Language Matching strategy for text-to-image synthesis, named VLMGAN*, which introduces a dual vision-language matching mechanism to strengthen the image quality and semantic consistency. The dual vision-language matching mechanism considers textual-visual matching between the generated image and the corresponding text description, and visual-visual consistent constraints between the synthesized image and the real image. 
Given a specific text description, VLMGAN* firstly encodes it into textual features and then feeds them to a dual vision-language matching-based generative model to synthesize a photo-realistic and textual semantic consistent image. 
Besides, the popular evaluation metrics for text-to-image synthesis are borrowed from simple image generation, which mainly evaluate the reality and diversity of the synthesized images. Therefore, we introduce a metric named Vision-Language Matching Score (VLMS) to evaluate the performance of text-to-image synthesis which can consider both the image quality and the semantic consistency between synthesized image and the description.  
The proposed dual multi-level vision-language matching strategy can be applied to other text-to-image synthesis methods.  We implement this strategy  on two popular baselines, which are marked with ${\text{VLMGAN}_{+\text{AttnGAN}}}$ and ${\text{VLMGAN}_{+\text{DFGAN}}}$ . The experimental results on two widely-used datasets show that the model achieves significant improvements over other state-of-the-art methods.
\end{abstract}


%
\IEEEpeerreviewmaketitle

\section{Introduction}

\begin{figure}
   \begin{center}
   \centering
     \includegraphics[scale=.7]{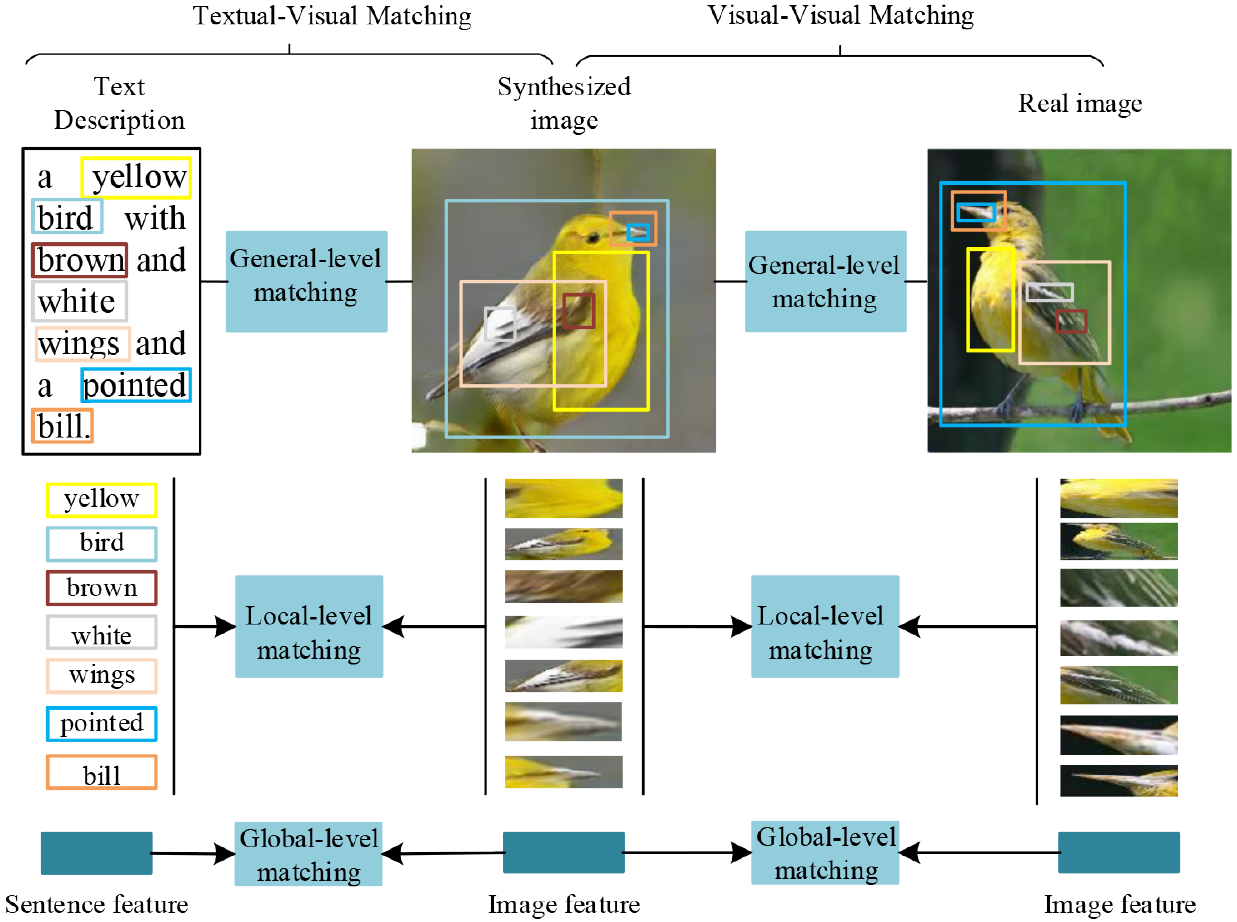}
     \caption{Illustration of the dual multi-level Vision-Language Matching that presents the basic idea of learning text-to-image synthesis by strengthening the semantic and visual matching of generated image with the real image and the corresponding text description. The bounding boxes indicate different image local-level feature, which are extracted by pre-trained model.}
  \label{figure1}
  \end{center}
\end{figure}

Photo-realistic image synthesis \cite{image_generation, image_synthesis} has drawn lots of attention in recent years, which has many potential applications, such as art design, computer graphics, and so on. Deep Neural Networks (DNNs) is powerful model for image related tasks, such as image generation \cite{imagecaption,image_synthesis} and image encoding \cite{ref_jpeg}. Remarkably, Generative Adversarial Networks (GANs)  ~\cite{GAN} is a milestone for image synthesis.
GANs-based methods have achieved incredible results on various image synthesis tasks, especially in 
high-resolution image generation ~\cite{biggan},
image style transfer ~\cite{cycgan} , and text-to-image synthesis ~\cite{AttnGAN,CFA-HAGAN,DMGAN,GAN-INT-CLS,MirrorGAN, cookgan,PPAN}. Among them, text-to-image synthesis is one key research sub-direction of Generative Adversarial networks. To be specific, text-to-image synthesis aims to generate photo-realistic and text-consistent image based on a specific text description.   
Image synthesis conditioned on natural language description has become an attractive direction, which presents great potential application in practical. For this task,  a large mount of approaches ~\cite{GAN-INT-CLS, AttnGAN, Bridge-GAN, CFA-HAGAN, ControlGAN, DFGAN, DMGAN, HAGAN, ktgan, SDGAN, StackGAN,StackGAN++} have been proposed to deal with this issue. 
The technique direction of text-to-image contains traditional DNNs and GANs. The former \cite{DRAW} adopts a DNNs to recover the image like other image reconstructions \cite{ref_sr, image_synthesis}.  
Generally, GANs-based text-to-image synthesis has two branches of techniques, one-stage framework and multi-stage framework.
One-stage framework follows the conventional GANs framework, which contains only one generator and one discriminator, such as GAN-INT-CLS \cite{GAN-INT-CLS}, DFGAN \cite{DFGAN}. For example, GAN-INT-CLS ~\cite{GAN-INT-CLS} concatenates the text embedding vector with a random noise vector and then feeds it into the generator to synthesis text-conditioned image. Multi-stage framework for text-to-image synthesis consists of multiple generators and discriminators, which are stacked in a pipeline, such as StackGAN ~\cite{StackGAN}, StackGAN++ \cite{StackGAN++}, DMGAN \cite{DMGAN}, MA-GAN \cite{MAGAN}. 
Compared with one-stage framework, multi-stage 
framework ~\cite{StackGAN, StackGAN++,  DMGAN} is also a popular solution for text-to-image 
synthesis, which firstly generates relatively blur and low-resolution images and then refines them to photo-realistic 
and high-resolution images. Since attention mechanism ~\cite{Attention} shows excellent performance in various tasks such as language translation, combining attention 
mechanism with multi-stage GANs ~\cite{AttnGAN} shows excellent performance in text-to-image synthesis. AttnGAN becomes an popular baseline and many researches follow their work, such as ~\cite{Bridge-GAN} , ~\cite{DMGAN}, CFA-HAGAN \cite{CFA-HAGAN}. Cheng et al. proposed CFA-HAGAN ~\cite{CFA-HAGAN} for text-to-image synthesis, which contains cross-modal attention and self-attention in the generation framework. SAMGAN ~\cite{SAMGAN} also adopts self-attention to support text-to-image synthesis in GAN.

Text-to-image synthesis is significantly different from simple image synthesis, which contains two challenges, visual reality and textual-visual semantic consistency.
The visual reality of image synthesis has been fully studied with the development of generative models, especially GANs \cite{GAN}. Some approaches \cite{image_generation,biggan} can generate highly realistic images, which are even difficult for our human to distinguish.
The visual-textual semantic consistency is the key challenge for text-to-image synthesis on account of the variegated text description. 
Although many approaches have ability in synthesizing relatively fine-grained and realistic images, especially for simple datasets such as CUB dataset ~\cite{CUB}, they rarely concentrate on the multi-level semantic consistency between the generated images and the corresponding texts.  
Recent approaches can synthesize relatively realistic images while they may fail to generate images that are semantically consistent with the text.
For example, for a description ``small bird, with white breast, 
red head and black wings and back'', the images synthesized by DMGAN ~\cite{DMGAN} and AttnGAN ~\cite{AttnGAN} 
do not identify with the description especially ``red head" as well as the ground truth image although they look realistic, as shown in the fourth column of Figure ~\ref{figure4}. 
Therefore, under the condition of photo-reality of image, text-to-image synthesis should focus on both the textual-visual matching and visual-visual matching simultaneously. Textual-visual matching can keep the image content consistent with the text description. The visual-visual matching consider the image quality and semantics of image content.

Besides, how to fairly evaluate the performance of the text-to-image synthesis is a significant issue that needs to be dealt with. As mentioned before, text-to-image synthesis aims at generating both realistic and semantic consistent image. Therefore, the evaluation metric should also include the two aspects, visual reality and textual-visual consistence. The popular evaluation metrics (IS ~\cite{IS} and FID \cite{FID}) mainly consider the visual reality, which are widely-used in image generation and image in-painting. To be specific, the IS calculates the KL divergence between the generated data and the original data, which are extracted by the pre-trained Inception-v3 model. The FID measures the Fréchet Distance between synthetic data and real data, which is extracted by the pre-trained Inception-v3 model. As is known, the accepted Inception-V3 is pre-trained by classification task on ImageNet \cite{ImageNet} dataset, which contains up to several million images. Besides, most images only contain one object, which is usually located in the center. Therefore, this a gap between the distribution of ImageNet and the chosen datasets.
Besides, the text-to-image synthesis is a pairing translation task, which should consider the synthesized image's quality and the semantic consistency with the text description. However, both FID and IS only consider the synthesized images while ignoring the text description. Thus, text-to-image synthesis needs a new evaluation metric which takes the consistency between the description and the synthesized image to make up for deficiencies of FID and IS. Meanwhi, the text-to-image task is suitable for a pairing evaluation metric that can take the two aspect into account.


Motivated by the mentioned observations, we aim to synthesize highly semantically consistent and photo-realistic images from the perspective of dual visual-textual matching and evaluate them under both visual reality and visual-textual semantic consistency.
To this end, we first propose a novel vision-language matching model (VLM) that can effectively explore the similarity between image and text based on metric learning. Then, we view the proposed VLM as an additional constraint block and insert it into a multi-stage GANs-based text-to-image synthesis framework. Besides, multi-level matching between the synthesized image and the real image is also considered. 
Figure ~\ref{figure1} shows the basic idea of the proposed method, which aims to strengthen both the textual-visual matching between the synthesized image 
and the text description and the visual-visual consistency between the synthesized image and the real photo-realistic image. 

According to the basic idea, we propose a novel metric for evaluating text-to-image synthesis performance in another view, 
called Vision-Language Matching Score (VLMS). As mentioned before, text-to-image synthesis focuses on both the image quality and the semantic consistency between the image contents and text description.
The proposed VLMS is obtained by a pre-trained Vision-Language Matching model, which is trained by ground-truth image-text pair data with metric learning. Experiments and analyses show that this metric can consider both the visual reality and textual-visual consistency. 

The critical contributions of our VLMGAN approach are listed as follows. 

\begin{enumerate}
   \item We design a dual semantically consistent text-to-image synthesis framework that can strengthen the textual-visual consistency between the visual content and textual 
   description and the visual-visual consistency between the synthesized image and real image. This mechanism is plug and play, which can be applied to any other text-to-image task.
   \item We propose a novel multi-level Vision-Language Matching model to learn the similarity between
   image and text, which can consider the global-level matching, fine-grained local-level matching, 
   and general-level matching. This model is optimized by metric learning, which can push the image and text into interpretable representation space. 
   \item A novel  evaluation metric (Vision-Language Matching Score, VLMS) is introduced in text-to-image synthesis to evaluate the performance. The VLMS considers both the visual reality of generated image and the semantic consistency between the generated image and the text description.
\end{enumerate}

We evaluate the proposed dual vision-language matching strategy on two baselines, AttnGAN and DFGAN. The experiments are conducted on two widely-used datasets, Caltech-UCSD Birds 200 (CUB) and Microsoft Common Objects in Context (MSCOCO). The synthesized image quality is evaluated under the popular metrics, IS, FID, 
R-precision, and the proposed VLMS. The experimental results demonstrate that the proposed method achieves impressive performance improvement over previous methods.
The rest of this paper is organized as follows: Section \ref{related_work}
reviews the related works of text-to-image synthesis briefly.
The methodology of our proposed VLMGAN* is introduced
in Section \ref{method}. Then we present and analyze the experimental
results in Section \ref{experiments}. Finally, we introduce the conclusion of this paper and the future work of our study in Section \ref{conclusion}.

\section{Related Work}
\label{related_work}
\subsection{Text-to-Image synthesis}
Text-to-Image synthesis aims at generating photo-realistic image that is also highly semantic consistent with the text description. 
For this task, many researchers present various kinds of technical solutions, such as variational inference ~\cite{PCNN}, conditional PiexlCNN ~\cite{reed2017}, and conditional generative adversarial networks ~\cite{GAN-INT-CLS}. Mansimov et al. ~\cite{mansimov2015generating} introduce a soft attention mechanism into DRAW \cite{DRAW} method to align the text description and the synthetic image.  Originally, Reed et al. ~\cite{reed2017} propose a conditional PixelCNN ~\cite{PCNN} 
based approach to synthesize a photo-realistic image from the text description.  Generative Adversarial Networks ~\cite{GAN} have shown surprising performance in various generative tasks, 
specifically in image synthesis ~\cite{biggan} via adversarial learning. For text-to-image synthesis, GAN also becomes the most popular research direction, such as \cite{GAN-INT-CLS, AttnGAN, Bridge-GAN, CFA-HAGAN, DMGAN, MirrorGAN, StackGAN, StackGAN++, Obj-GAN, HDGAN, DFGAN} and so on. For instance, Reed et al. ~\cite{GAN-INT-CLS} firstly decompose text-to-image synthesis into two subtasks, encoding the text description to a unique representation and then synthesizing images conditioned on this vector by generative adversarial networks, called GAN-INT-CLS. An improved approach, named Generative Adversarial What-Where Network (GAWWN) ~\cite{GAWWN}, can focus on the location where objects should be drawn. 
Nguyen et al. ~\cite{PPGN} propose Plug and Play Generative Networks (PPGN) to generate images by interpreting activation maximization. 

The multi-stage text-to-image synthesis framework, firstly introduced in StackGAN ~\cite{StackGAN}, shows remarkable superiority comparing to the one-stage strategy. This critical thought is widely accepted and applied by many kinds of research \cite{StackGAN++, AttnGAN, DMGAN, Bridge-GAN, CFA-HAGAN}, which gradually improves the image resolution and quality. Specifically, StackGAN \cite{StackGAN} adopts two-stage GANs to synthesize high-resolution images gradually: the first generator synthesizes 64x64 pixel images, and then the second generator refines it to 128x128 resolution. 
Based on StackGAN, a more advanced version StackGAN++ ~\cite{StackGAN++} is proposed, which has three generators. Besides, HDGAN ~\cite{HDGAN} introduces a patch-wise adversarial loss into 
multi-stage generative framework.

 For synthesizing image conditioned on fine-grained word-level textual features, AttnGAN ~\cite{AttnGAN} adopts the attention mechanism into a multi-stage generative framework. MirrorGAN ~\cite{MirrorGAN} introduces a mirror procedure in the text-to-image task, which firstly conducts text-to-image generation and then re-describes 
the synthesized image. Zhu et al. ~\cite{DMGAN} propose a Dynamic Memory Generative Adversarial Network (DM-GAN), which can refine the generated image by a dynamic memory block. SDGAN ~\cite{SDGAN} adopts conditional batch normalization to reinforce the text highly relevant elements in the image features. DFGAN ~\cite{DFGAN} fuses the text information into the hidden visual feature by a novel deep visual-textual fusion block in the image synthesis procedure. It should be noted that DFGAN adopts one-stage  framework rather than multi-stage framework for text-to-image synthesis. ControlGAN ~\cite{ControlGAN} introduces spatial and channel-wise attentive mechanism and perceptual loss to synthesize high-quality image. LeicaGAN ~\cite{LeicaGAN} introduces multiple prior knowledge to enforce semantic consistency. RiFeGAN ~\cite{rifegan} learns rich feature for text-to-image generation from prior knowledge.

Obj-GAN ~\cite{Obj-GAN} can focus on synthesizing object aware images by object-driven attentive generative network. Besides, its discriminator adopt Fast R-CNN with a binary cross-entropy loss to discriminate the object information of each bounding box. Tobias et al. ~\cite{OP-GAN} introduce OPGAN to focus  on individual objects in generating. CPGAN ~\cite{cpgan} adopts Yolo-V3 ~\cite{yolov3} to design an image content-aware discriminator in the text-to-image framework, which shows remarkable performance. Dong et al. ~\cite{UGAN} propose a text-to-image synthesis model in an unsupervised learning manner, which does not rely on the human-labeled data. Wang et al. ~\cite{end2end} propose an end-to-end framework for text-to-image. Recently, DALLE ~\cite{dalle} shows amazing performance on generating image from text with the help of pre-training on tons of data, which also verifies the importance of data volume. However, the images synthesized by DALLE \cite{dalle} are usually cartoon style, which may be due to the pre-training dataset containing large mount of cartoon picture.

\subsection{Image-text Matching}
Cross-modal understanding is an attractive but challenging task, which includes cross-modal retrieval \cite{dsran,tcsvt}, image captioning \cite{imagecaption} and semantic grounding \cite{ref_videoseg}.  Specifically, image-text matching plays a key role in cross-modal retrieval.
An image-text matching model aims to project the visual image and textual description into a semantic shared space using contrastive learning \cite{moco, tan2022unified, chen2022utc}. The heterogeneity gap between various type of data can be bridged by the mapped space. Specifically, visual and textual features are encoded separately into the same subspace, where the similarity values can be directly calculated. Methods can be divided into three kinds: global matching methods, regional matching methods, and multi-level matching methods.

For global matching methods, images and texts are encoded in a global way either with a CNN ~\cite{resnet}, or an LSTM ~\cite{lstm}. The visual features and textual features are then embedded into the subspace, where their global similarity can be computed and optimized by a triplet-ranking loss ~\cite{vse++}. Since CNNs for image feature extracting are pre-trained on ImageNet ~\cite{ImageNet}, for text feature extracting, a pre-trained BERT ~\cite{bert} can be used for more refined features, as did in COOKIE ~\cite{cookie}. 

However, these methods fail to match the concrete objects in the raw image and words in the sentence, which can be solved by regional matching methods. Thus SCAN ~\cite{scan} uses a pre-trained Faster RCNN ~\cite{faster} to detect the concrete objects and designs a stacked cross attention mechanism to align objects and words. 
Further, VSRN ~\cite{vsrn} adopts graph convolution ~\cite{gcn} to learn the regional relations corresponding to the textual relations. 
With cross-modal pre-training and transformer-based encoders, the similarity score of image and sentence can be directly learned instead of distance calculation, as did in Uniter ~\cite{uniter}.  

Considering both regional and global cross-modal matching, multi-level matching methods learn the object-word alignment and global semantic alignment simultaneously. 
GSLS ~\cite{gsls} designs a multi-path structure to get both global and local similarities. 
CRAN ~\cite{cran} designs a multi-path structure for learning the global, local, and relational alignment at the same time. Wen et al. ~\cite{dsran} utilized GAT ~\cite{gat} to learn dual relations of image objects and backgrounds in alignment with phrases in sentences. To calculate the similarity between the generated image and the original sentence more comprehensively, we design a multi-level matching model VLM in our method.

\section{Visual-Language Matching GAN}
\label{method}
\begin{figure*}
   \begin{center}
   \centering
     \includegraphics[width=1\linewidth]{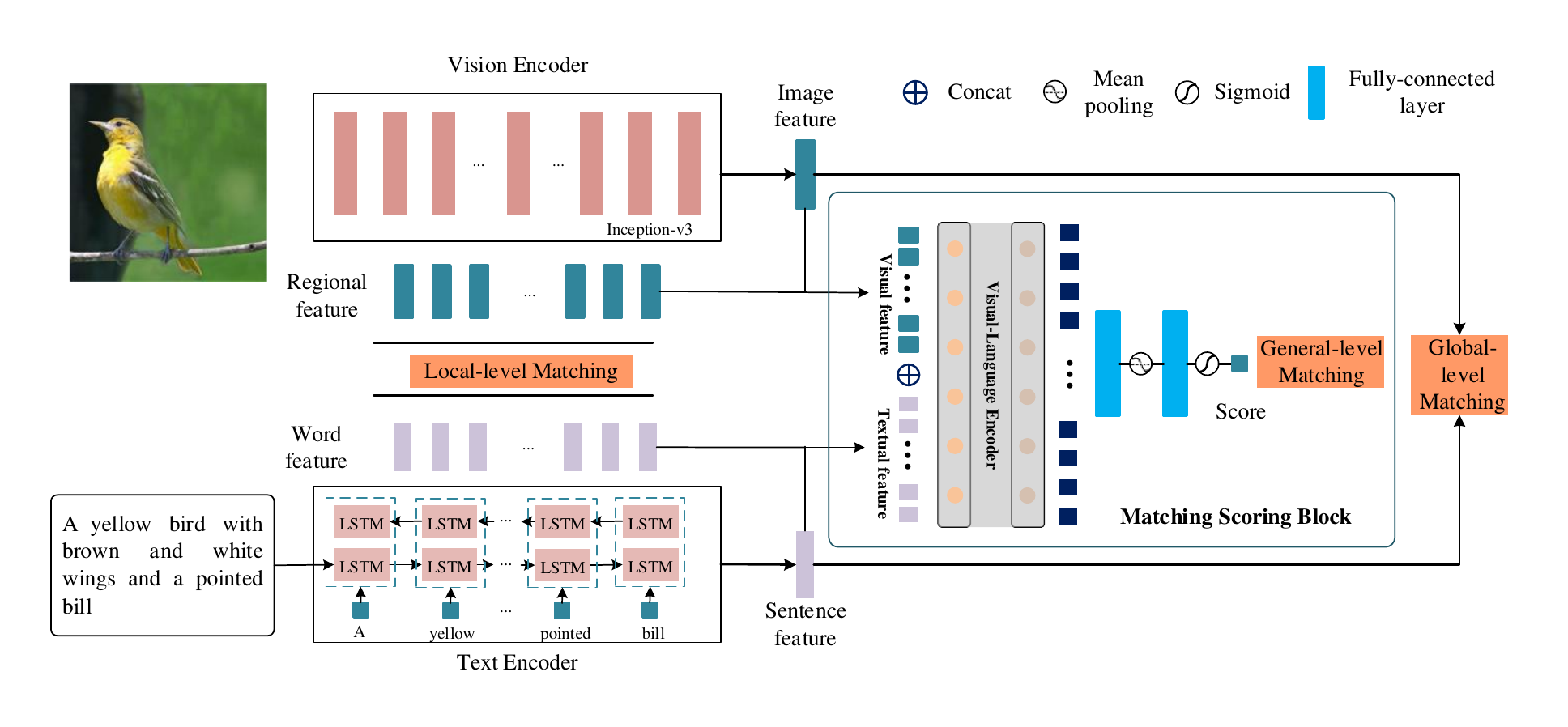}
  \caption{Illustration of the multi-level vision-language matching model. The Visual-Language matching model contains three sub-models: Vision Encoder, Text Encoder and Matching Scoring Block.}
  \label{figure2}
  \end{center}
\end{figure*}

\subsection{Vision-Language Matching Model}

The vision-language matching (VLM) model learns the multi-level similarity of text and image modality, including local-level matching, global-level matching, and general-level matching. The architecture of the proposed VLM model is shown in Figure \ref{figure2}. 
The VLM model contains three sub-models: Vision Encoder, Textual Encoder and Matching Scoring Block (MSB). The Vision Encoder and Text Encoder aim at embedding the image and text into semantic aligned vectors, which is a key process for connecting the domains of vision and language.  
For a fair comparison, we adopt the same backbone (Inception-v3  ~\cite{Inception-v3} for image and Long Short-Term Memory  ~\cite{lstm} for text) with 
DAMSM ~\cite{AttnGAN} to extract the semantic features. Inception-v3 \cite{Inception-v3} is a widely-used model for visual feature extraction.
LSTM \cite{lstm} can solve long distance memory problem, which is widely-used in natural language processing.
The proposed MSB plays a role of generating a matching score for the image and text by a transformer encoder.

\textbf{Text Encoder.} 
Embedding text language, the simplest approach is adopting bag-of-words (BOW) model, which is widely-used in many tasks, such as cross-modal retrieval ~\cite{GRL}. However, BOW does not consider the semantic context of the text description, which is gradually replaced by learning-based model, such as LSTM \cite{lstm} and Skip-gram \cite{skip}. 
For the $i$-word of a sentence, an embedding layer embed it into a semantic vector ${w_i}$ and then feed into the LSTM. Specifically, for a text, the word feature is denoted by the hidden states, and the sentence feature is represented by the last hidden state. 
\begin{equation}
   \begin{aligned}
       {  \varphi, \overline { \varphi } = F_{\text{Text-Encoder}}(w_1, w_2, ..., w_n)},
    \end{aligned}
\end{equation}
where ${\varphi}$  (matrix size:${256 \times T_0}$, ${T_0 }$ is the sentence length) is the word feature matrix and ${ \overline { \varphi }}$ is the sentence feature.

\textbf{Vision Encoder.}
The visual feature is extracted by a Convolutional Neural Network, named Inception-v3 \cite{Inception-v3}. Following previous works \cite{AttnGAN}, the intermediate features of CNNs can present the local regional feature of an image, while the feature of the last layer is the global feature of an image.
The Inception-v3 ~\cite{Inception-v3} model is pre-trained on ImageNet ~\cite{ImageNet}. The local-regional features ${f}$ (${768\times 17 \times 17}$) are denoted by the output of the ${mixed\_6e}$ 
layer and the global features ${\overline{f}}$ (${2048 \times 1 }$) are represented by the ${Mixed\_7b}$ layer. With reshaping and linear projection, ${\Phi}$ (${768\times 289}$) denotes the 
local-regional feature and ${\overline { \phi }}$ represents the global image feature. The projection is shown as follows,

\begin{equation}
   \begin{aligned}
       {  \phi = F_{1\times1 conv}(f),   \overline { \phi } = W \overline { f },  }
    \end{aligned}
\end{equation}
 where ${\phi \in \mathbb{R}^{D\times289}}$ and  ${\overline{\phi} \in \mathbb{R}^{D}}$. ${D}$ is the dimension of visual and textual feature, which is equal to 256. It should be noted that only the newly added layers are trainable. 
 
 The whole process of visual feature extraction can be presented by the following formula,
 \begin{equation}
   \begin{aligned}
       {  \phi, \overline { \phi }= F_{\text{Vision-Encoder}}(x)}
    \end{aligned}
\end{equation}
 
\textbf{Matching Scoring Block.} 
The vision-language matching scoring block aims at producing a general matching score to evaluate the matching degree between the image and the text. Transformer ~\cite{Attention} has shown promising performance in various tasks especially in vision-language understanding, such as Bert ~\cite{bert} and Uniter \cite{uniter}. Self-attention in Transformer \cite{Attention} can deeply explore the semantic relations between visual feature and textual feature. Therefore, we adopt this mechanism to learn the matching score between image and text. Calculating the general-level matching considers both global feature and local features of the image  and sentence feature and word feature of text. Specifically, the vision-language united feature is defined as
\begin{equation}
 \begin{aligned}
    { \psi = F_{cat}(\varphi, \overline{\varphi}, \phi, \overline{\phi}). }
  \end{aligned}
\end{equation}
Where $F_{cat}$ mean the concat operation.
Then, the united feature is feed into the Transformer-based vision-language encoder, as follows.
  
\begin{equation}
 \begin{aligned}
    { \hat{\psi} = F_{Transformer}(\psi) .}
  \end{aligned}
\end{equation}
After obtaining the visual-textual latent features, a fully connected layer is chosen to project the features into a hidden space.
\begin{equation}
 \begin{aligned}
    { \hat{\psi} = W_0\hat{\psi} + b_0, }
  \end{aligned}
\end{equation}
where ${W_0}$ and $b_0$ are the learn-able parameters of the fully connected layer. 
The final feature can be obtained by Mean pooling.
\begin{equation}
 \begin{aligned}
    { \overline{\psi} = F_{mean\_pooling}(\hat{\psi}) .}
  \end{aligned}
\end{equation}

Lastly, the vision-language matching score is calculated by a Sigmoid function after a Fully Connected layer, which projects the feature into a 1-dimensional value.   
\begin{equation}
 \begin{aligned}
    { Score = F_{Sigmoid}(W_1\overline{\psi} + b_1), }
  \end{aligned}
\end{equation}
where ${W_1}$ and $b_1$ are the learn-able parameters of the second fully connected layer. 
The whole process of learning the visual-language matching score can be denoted by the following formula,
\begin{equation}
 \begin{aligned}
      Score = F_{\text{MSB}}(\phi, \overline { \phi}, \varphi, \overline { \varphi }).
      \label{vlms_eq}
  \end{aligned}
\end{equation}

\textbf{Local-level matching.} 
The local-level matching considers the semantic consistence between the word features and image local-regional features. Given a specific image with 289 local regional features and a text description with 
${T_0}$ word features, the cosine similarity for all possible image region and word pairs are calculated by the following formula,
\begin{equation}
 \begin{aligned}
{s ( \phi_{{i}},\varphi_{{j}} ) =\frac{{\phi_{{i}}^{\text{T}} \varphi _{{j}}}}{{{ \left\Vert {\phi_{{i}}} \right\Vert }{ \left\Vert { \varphi _{{j}}} \right\Vert }}}, i \in [1,2,...,289], j \in [1,2,...,T_0]. }
  \end{aligned}
\end{equation}
Here, ${s ( \phi_{{i}},\varphi_{{j}}})$ is the similarity between the ${i}$-th image region and ${j}$-th word. We adopt ${S}$ be the similarity matrix between word features and image local features. We adopt the popular attention mechanism \cite{Attention} to learn the fine-grained similarity. The word context with respect to each image region is calculated by a weight sum of image visual feature, as following.

\begin{equation}
 \begin{aligned}
{c_{{i}}={\mathop{ \sum }_{{j=0}}^{{288}}{ \alpha _{{ij}} \phi _{{j}}}},}
  \end{aligned}
\end{equation}
where
\begin{equation}
 \begin{aligned}
{\alpha _{{ij}}=\frac{{exp \left(  \gamma _{{1}}s_{{i,j}} \right) }}{{{\mathop{ \sum }_{{k=0}}^{{288}}{exp \left(  \gamma _{{1}}s_{{i,k}} \right) }}}}}.
  \end{aligned}
\end{equation}
Here, ${\gamma _1}$ is the in-versed temperature of the softmax function, set as 4.



Following minimum classification error formulation in speech recognition \cite{minimum}, the local-level matching score between the image and the text description is calculated by the LogSumExp pooling, as following, 
\begin{equation}
 \begin{aligned}
    {{S_{local} ( I,T ) =log \left( {\mathop{ \sum }\limits_{{i=1}}^{{T-1}}{exp \left(  \gamma _{{2}}S \left( c_{{i}}, \varphi _{{i}} \right)  \right) }} \right) }^{{\frac{{1}}{{ \gamma _{{2}}}}}},}
  \end{aligned}
\end{equation}
where ${S ( c_{{i}},\varphi_{{i}} )}$ is the matching score between the ${i}$-th word and the ${i}$-th region-context, calculated by cosine similarity, ${\gamma _{{2}}}$ is an adjusting factor, set as 5.
\begin{equation}
 \begin{aligned}
{S ( c_{{i}},\varphi_{{i}} ) =\frac{{c_{{i}}^{\text{T}} \varphi _{{i}}}}{{{ \left\Vert {c_{{i}}} \right\Vert }{ \left\Vert { \varphi _{{i}}} \right\Vert }}}. }
  \end{aligned}
\end{equation}

\textbf{Global-level matching.}
The global-level matching considers the visual global feature and the global textual feature.
Similarly, for global visual feature ${ \overline {  \phi}}$  and the sentence feature ${\overline {  \varphi}}$ , the matching score is directly calculated by the cosine similarity,
\begin{equation}
 \begin{aligned}
    { S_{global} \left( I,T \right) =\frac{{ \overline { \varphi }^{\text{T}} \overline { \phi }}}{{{ \Vert { \overline { \varphi }} \Vert }{ \Vert { \overline { \phi }} \Vert }}}. }
  \end{aligned}
\end{equation}

\textbf{General-level matching.}
The general-level matching score is produced by the pre-trained vision-language Matching Scoring Block. As follows, 
\begin{equation}
 \begin{aligned}
    { S_{general}(I,T) = F_{\text{MSB}}(\phi, \overline { \phi}, \varphi, \overline { \varphi }). }
  \end{aligned}
\end{equation}

\textbf{Objective function.} Triplet loss is a popular ranking objective for matching task, which is widely-used in image-text matching \cite{dsran, vsrn,uniter}. After obtaining the matching scores of three levels,  the hinge-based triplet ranking loss \cite{vse++} is adopted to optimize the vision-language matching model. The optimization loss for general-level matching is defined below.
\begin{equation}
\begin{aligned}
\mathcal{L}_{general}=&[\alpha+S_{general}(I^{'}, T)-S_{general}(I, T)]_+ + \\
&[\alpha+S_{general}(I, T^{'})-S_{general}(I, T)]_+, \label{eq11} 
\end{aligned}
\end{equation}
where ${S_{general}(I^{'}, T)}$ and ${S_{general}(I, T^{'})}$ are the general-level matching scores of un-pairing image-text instance,  ${S_{general}(I, T)}$ are the general-level matching scores of pairing image-text instances, and ${\alpha}$ is a margin. In our experiments, ${\alpha}$ is set as 0.2. If the image and the text are closer to one another in the joint embedding space than any negatives pairs, by the margin ${\alpha}$, the hinge loss is zero. If we substitute the general-level matching score by local-level and global-level matching score, we can obtain the loss of ${\mathcal{L}_{local}}$ and ${\mathcal{L}_{global}}$. 

Finally, the overall loss function of VLM model is defined as
\begin{equation}
\begin{aligned}
\mathcal{L}_{VLM}= \mathcal{L}_{local} +\mathcal{L}_{global}+\mathcal{L}_{general}.
\end{aligned}
\end{equation}

\textbf{Optimization.} The Optimization of VLM  contains two types. The first type is training for supervising text-to-image synthesis, which is optimizing on the training dataset. The second type is optimizing the Matching Scoring Block for obtaining the VLMS to evaluate the performance. This type is training on the whole dataset including the testing dataset. The optimizer is Adam and the learning rate is 0.0002. The training is stopped after 200 epochs.

\subsection{Dual Matching-driven Attentive GAN}
\begin{figure*}
   \begin{center}
   \centering
   \includegraphics[scale=0.66]{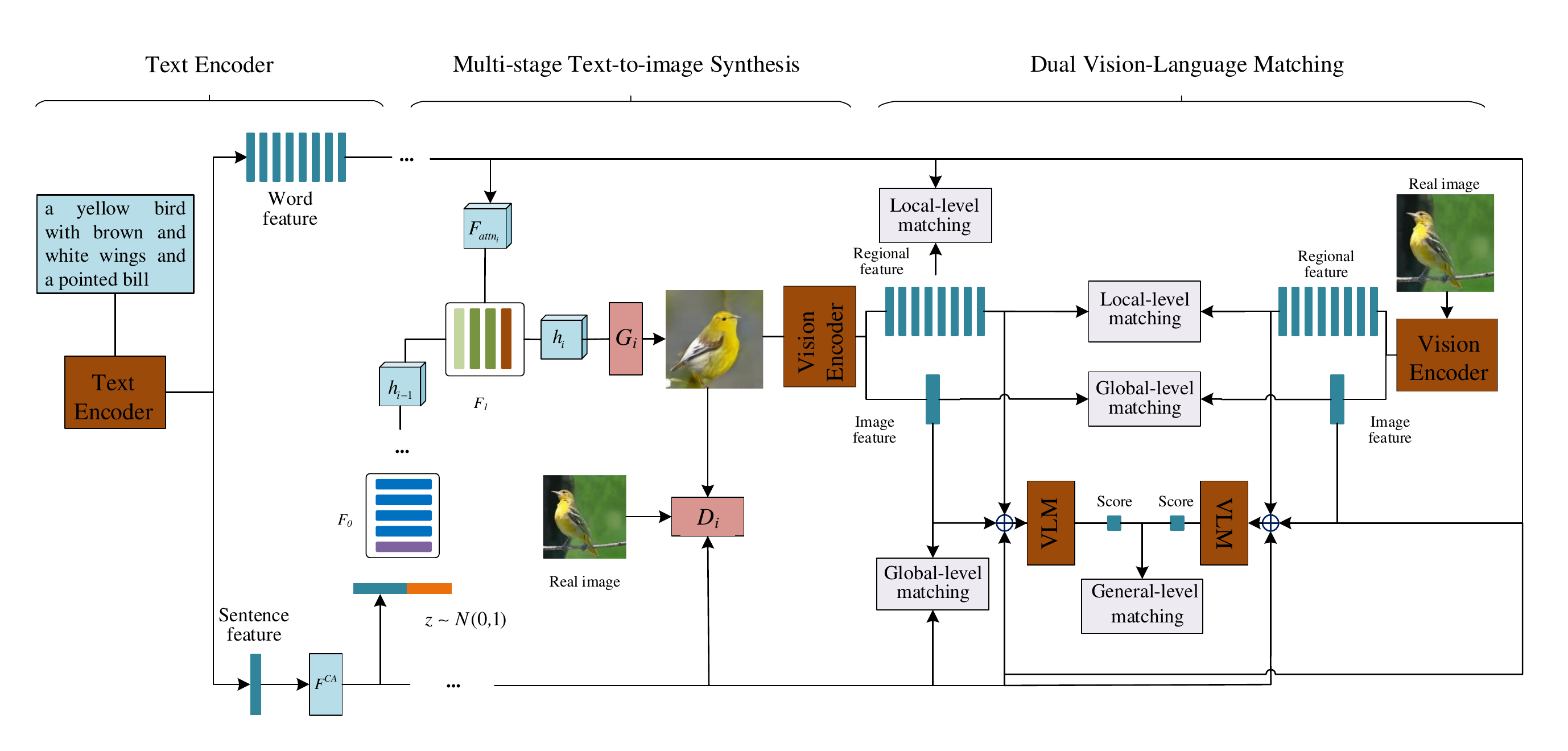}
   \caption{The text-to-image architecture of the proposed ${\text{VLMGAN}_{+\text{AttnGAN}}}$. The ${\text{VLMGAN}_{+\text{AttnGAN}}}$ also adopts the popular attentive multi-stage strategy to improve the image quality gradually. Besides, dual vision-language matching module provides semantic consistency supervision.}
  \label{figure3}
  \end{center}
\end{figure*}
For a fair comparison and better understanding, we adopt AttnGAN ~\cite{AttnGAN} as baseline, which is also chosen by many state-of-the-art methods due to its excellent performance, to implement the dual vision-language matching strategy.
As a example, the ${\text{VLMGAN}_{+\text{AttnGAN}}}$ 
The multi-stage text-to-image synthesis architecture is stacking three generative-adversarial blocks sequentially, as shown in the left part of Figure ~\ref{figure3}.  
Given a specific text description, the sentence feature ${\overline{\varphi}}$ and word features ${\varphi}$ are extracted by the text encoder in VLM model. The synthesized image 
can be obtained by the following procedures.
\begin{equation}
   \begin{aligned}
    { h_0=F_0{(F_{cat} ( z, F^{ca} (  \overline { \varphi } ) ))},} \\
    {h_{i}={F}_{i} ( h_{{i-1}},F_{{attn_i}} (h_{i-1}, \varphi )  ) 
  , i \in \{1, 2. \}} \\
    {\hat{\mathop{{x}}}_{{i}}=G_{{i}} ( h_{{i}} ),  i \in \{0,1,2.\} }
   \end{aligned} 
\end{equation}
where ${z \sim N(0, 1)}$ is a random noise vector, ${F^{ca}}$ is conditioning augmentation (CA) ~\cite{StackGAN} process, and
${F_{attn_i}}$ is attentive mechanism described in AttnGAN ~\cite{AttnGAN}. Due to the excellent performance of attentive mechanism, We adopt the word-level attentive mechanism ${F_{attn_i}}$ to fuse the word features into visual features. The ${F_{attn_i}}$ has two inputs, the word features $\varphi$ and previous hidden features $h$. Mathematically, the word-visual fusion context can be defined as follows,
\begin{equation}
   \begin{aligned}
  {h_{i} =F_{attn_{i}} (\varphi ,h_{i-1} ) =  ( c_{0},c_{1}, \cdots ,c_{N} )  \in  \mathbb{R}^{\hat{D} \times N}, }
   \end{aligned}
\end{equation}
where
\begin{equation}
   \begin{aligned}
      {c_{j}={\mathop{ \sum }\limits_{{i=0}}^{{T_0-1}}{ \alpha _{{j,i}}\varphi_{i}}.}}
   \end{aligned}
\end{equation}
Here, ${\alpha _{{j,i}}}$ is the attention weight between the ${j}$-th image region and the ${i}$-th word feature, as follows. 
\begin{equation}
   \begin{aligned}
  {{ \alpha _{{j,i}}=\frac{{exp \left( {s ^ {\prime} }_{j,i} \right) }}{{{\mathop{ \sum }^{{T-1}}_{{k=0}}{exp \left( {s^{\prime} }_{j,k} \right) }}}}.}}
   \end{aligned}
\end{equation}
where ${s ^ {\prime}}_{j,i} = {h_j}^{\text{T}}\varphi_{i}$ and  ${s^ {\prime}}_{j,i} $ indicates the similarity between the $j$-th feature in hidden feature $h$ and the $i$-th word feature.

\textbf{Objective function.} Each stage contains a generator and discriminator, which are optimized alternately by a generative loss and a discriminating loss. Specifically, 
the generative loss is
\begin{equation}
\small
   \begin{aligned}
    {\mathcal{L}_{G_i} =-\frac{{1}}{{2}} E_{\hat{{x}}_i \sim P_{G_i}} [ log ( D_i ( \hat{x}_i )  ) ] -\frac{{1}}{{2}}E_{{\hat{x}}_i \sim P_{{G_i}}} [ log (D_i ( \hat{x}_i, \overline {\varphi}))],}
   \end{aligned}
\end{equation}
where ${x_i}$ is the synthesized image. In this objective function, the former is the unconditional loss, which determines whether the synthesized image is real or fake.  The latter is 
the conditional loss, which determines whether the image contents are matched with the text description or not. The generative loss function forces the model synthesize photo-realistic and text semantic consistent images.

At the same time, the discriminator is designed to distinguish the generated image is both fake and semantic consistent or not. Therefore, the discriminating loss also consists of unconditional visual realism loss and conditional semantic consistent loss. Mathematically, it is defined 
 as follows.
\begin{equation}
 \scriptsize
 \begin{aligned}
   {\mathcal{L}_{{D_{{i}}}}=-\frac{{1}}{{2}}E_{{x_{{i}} \sim P_{{data}}}} [ log ( D_{{i}} ( x_{{i}} )  )  ]-  \frac{{1}}{{2}}E_{{\hat{{{x}}}_{{i}} \sim P_{{G_{{i}}}}}} [ log ( 1-D_{{i}} ( \hat{{{x}}}_{{i}} )  )  ] +} \\
   {-\frac{{1}}{{2}}E_{{x_{{i}} \sim P_{{data}}}} [ log ( D_{{i}} ( x_{{i}}, \overline { \varphi } )  )  ]  -\frac{{1}}{{2}}E_{{\hat{{{x}}}_{{i}} \sim P_{{G_{{i}}}}}} [ log ( 1-D_{{i}} ( \hat{{{x}}}_{{i}}, \overline { \varphi } )  )  ], }
  \label{ddd}
  \end{aligned}
\end{equation}
where ${P_{data}}$ is the real data distribution and ${P_{G_i}}$ is the generated image data distribution. The first two expressions are unconditioned losses, which focus on distinguishing the synthesized image is real or fake. The later two expressions are conditioned losses, which focus on distinguishing the synthesized image is consistent with the text description or not. 
However, only adopting the discriminator to force the synthesized image consistence is insufficient. As shown in Equation ~\ref{ddd}, the discriminator only considers the global text feature instead of both global feature and local feature. 

To obtain more semantic consistent image, we introduce an additional supervision part, dual multi-level vision-language matching module.
The dual multi-level vision-language matching module contains two parts, textual-visual matching and visual-visual matching. We introduce the loss function ${\mathcal{L}_{VLM}}$ of VLM model to strengthen the textual-visual consistency between the generated image and the corresponding text. Besides, the visual-visual consistency between the synthesized image and the real image should be considered in the objective function. 

\textbf{Visual-Visual Matching (VVM)} also contains three parts, local-level matching, global-level matching, and general-level matching. We adopt the vision encoder to extract the local-level features ${\phi_{fake}}$ and global-level feature ${\overline{\phi}_{fake}}$ of the synthesized image ${\hat{x}}$, as following.
\begin{equation}
   \begin{aligned}
       {  \phi_{fake}, \overline { \phi}_{fake} = F_{\text{Vision-Encoder}}(\hat{x})}
    \end{aligned}
\end{equation}

Analogously, the local-level features ${\phi_{real}}$ and global-level feature ${\overline{\phi}_{real}}$ of the real image ${{x}}$
\begin{equation}
   \begin{aligned}
       {  \overline{\phi}_{real}, \overline { \phi}_{real} = F_{\text{Vision-Encoder}}({x})}
    \end{aligned}
\end{equation}

The visual-visual global-level matching loss aims at maximizing the global matching score between the real image and the synthesized fake image. The loss function is 
 \begin{equation}
\begin{aligned}
\small
    &\mathcal{ L}_{VG} (\phi_{real}, \overline{\phi}_{fake})=\\
    &-\frac{1}{B}\sum_{i=1}^B log \frac{e^{(S(\overline{\phi}_{real}, \overline{\phi}_{fake}^{+})/\tau_0)}}{e^{(S(\overline{\phi}_{real}, \overline{\phi}_{fake}^{+})/\tau_0)} + \sum_{j=1}^{B-1}e^{ (S(\overline{\phi}_{real}, \overline{\phi}_{fake}^{-})/\tau_0)}},
\end{aligned}
\end{equation}
where $B$ is the batch size and $\tau_0$ is a hyperparameter (set as 0.07). $S(\overline{\phi}_{real}, \overline{\phi}_{fake}^{+})$ is the cosine similarity between the paired real image and synthesized image. $S(\overline{\phi}_{real}, \overline{\phi}_{fake}^{-})$ is the cosine similarity between the unpaired real image and synthesized image. The loss function of Vision-Vision global matching is the sum of fake-to-real and real-to-fake, as following.
 \begin{equation}
\begin{aligned}
\small
    &\mathcal{ L}_{VG} =\mathcal{ L}_{VG} (\overline{\phi}_{real}, \overline{\phi}_{fake}) + \mathcal{ L}_{VG} (\overline{\phi}_{fake},\overline{\phi}_{real}),
\end{aligned}
\end{equation}

The visual-visual local-level matching loss maximizes the similarity between the regional features of real image and synthesized image.  
\begin{equation}
\begin{aligned}
\small
    &\mathcal{ L}_{VL} (\phi_{real}, {\phi}_{fake})=\\
    &-\frac{1}{B}\sum_{i=1}^B log \frac{e^{(S({\phi}_{real}, {\phi}_{fake}^{+})/\tau_0)}}{e^{(S({\phi}_{real}, {\phi}_{fake}^{+})/\tau_0)} + \sum_{j=1}^{B-1}e^{ (S({\phi}_{real}, {\phi}_{fake}^{-})/\tau_0)}},
\end{aligned}
\end{equation}
where  $S({\phi}_{real}, {\phi}_{fake})$ is calculated by the following formula.

\begin{equation}
 \begin{aligned}
    {{S({\phi}_{real}, {\phi}_{fake}) =log \left( {\mathop{ \sum }\limits_{{i=1}}^{{N-1}}{exp \left(  \gamma _{{3}}S \left(\varphi_{{i}}^{real}, \varphi_{{i}}^{fake} \right)  \right) }} \right) }^{{\frac{{1}}{{ \gamma _{{3}}}}}},}
  \end{aligned}
\end{equation}
Here $S(\cdot)$ is cosine similarity calculation formula and $\gamma _{{3}}$ is set as 5. 
The loss function of visual-visual local-level matching is the sum of two parts, as following.
 \begin{equation}
\begin{aligned}
\small
    &\mathcal{ L}_{VL} =\mathcal{ L}_{VL} (\phi_{real}, {\phi}_{fake}) + \mathcal{ L}_{VL} ({\phi}_{fake},\phi_{real}),
\end{aligned}
\end{equation}

The visual-visual general-level matching loss is calculated by the pre-trained MSB model, as following.
\begin{equation}
   \small
   \begin{aligned}
    {\mathcal{L}_{VGEN} = \parallel F_{\text{MSB}}(\phi_{real}, \overline { \phi}_{real}, \varphi, \overline { \varphi }) - F_{\text{MSB}}(\phi_{fake}, \overline { \phi}_{fake}, \varphi, \overline { \varphi }) \parallel^2 _2.}
   \end{aligned}
\end{equation}

The overall loss function of Visual-Visual Matching ${\mathcal{L}_{VVM}}$ is the sum of local-level matching ${\mathcal{L}_{VL}}$, global-level matching ${\mathcal{L}_{VG}}$, and general-level matching ${\mathcal{L}_{VGEN}}$, as following. 

\begin{equation}
   \begin{aligned}
    {\mathcal{L}_{VVM} = \mathcal{L}_{VG} + \mathcal{L}_{VL} + \mathcal{L}_{VGEN}.}
   \end{aligned}
\end{equation}

\textbf{Textual-Visual Matching (TVM)} is presented in Section Vision-Language Matching Model. In text-to-image model, we feed the synthesized image ${\hat{x}}$ in the VLM model to calculate the textual-visual matching loss ${\mathcal{L}_{VLM}}$.

The adversarial generative loss, the textual-visual matching loss and the visual-visual loss are combined in the general loss function. Therefore, the objective function of the overall generative model is defined as
\begin{equation}
   \begin{aligned}
    {\mathcal{L}_{G} = \mathop{ \sum }\limits_{{i=0}}^{{2}}\mathcal{L}_{G_i}  +\lambda_1\mathcal{L}_{VVM} + \lambda_2 \mathcal{L}_{VLM}.}
   \end{aligned}
\end{equation}
where $\lambda_1$ and $\lambda_2$ are two balancing factors.They are set as 5 for ${\text{VLMGAN}_{+\text{AttnGAN}}}$ and set as as 0.25 for ${\text{VLMGAN}_{+\text{DFGAN}}}$.
In the training process, the three generators are optimized simultaneously and the three discriminators are optimized independently. The parameters of vision encoder, text encoder, and matching scoring block are not trainable.

\section{Experiments}
\label{experiments}
In this section, extensive experiments are conducted to verify the effectiveness of the proposed method. 
We firstly introduce the experimental settings and then show the quantitative
and qualitative evaluation results. Lastly, we present ablation studies and further discussions.

\subsection{Datasets and Evaluation Metrics}
\textbf{Datasets.} Two widely-used benchmarks, CUB ~\cite{CUB}, and MSCOCO ~\cite{MSCOCO}, are adopted to demonstrate the capability of the proposed method. The CUB contains 11,788 images belonging to 200 categories, which is divided into two sub-datasets, 8,855 images for training and the remaining 2,933 for testing. For each image, there are ten textual descriptions. The MSCOCO dataset is a larger and more challenging benchmark. It contains 120k images, and each image is described by five texts. We split them into a training set with about 80k images and a testing set with 40k images. The dataset settings are the same as previous works. 

\textbf{Evaluation Metric.} We quantify the effectiveness of the proposed method in terms of Fréchet Inception Distance (FID) ~\cite{FID}, Inception Score (IS) ~\cite{IS}, 
R-precision ~\cite{AttnGAN}, and the proposed VLMS. 

The IS calculates the Kullback-Leibler (KL) divergence between the class distribution of original image and the class distribution of generated image. The class distribution is calculated by the pre-trained Inception-v3 model. The higher IS suggests that the synthesized images are more realistic and more confident to a specific class. The Inception Score is calculated by the following formula:
\begin{equation}
 \begin{aligned}
{\text{IS}=exp ( E_{{X \sim P_{{G}}}} [ D_{KL} ( P_{{Y | X  }} ( y | x )  ) |  | P_{{Y}} ( y )  ]) ,  }
  \end{aligned}
\end{equation}
where ${x}$ is the generated fake image, and ${y}$ is its corresponding semantic label predicted by the pre-trained Inception-v3 model. The distribution ${p \left( y \left| x \right) \right. }$ denotes the probability distribution of image x belongs to a specific category ${y}$ , and ${p \left( y \right)}$  is the probability distribution of predicted class. 

The FID measures the Fréchet Distance between global semantic feature of synthesized image and real image, which are extracted by the pre-trained Inception-v3 model. Thus, lower value of FID ndicates that the synthesized images are close to the original images. Lower value is better, vice versa. The calculation of FID is as follows.
\begin{equation}
 \begin{aligned}
    {\text{FID}={ \left\Vert {m-m_{r}} \right\Vert }_{{2}}^{{2}}+Tr \left( C+C_{r}-2 \left( CC_{r} \left) ^{{\frac{{1}}{{2}}}} \right) ,\right. \right.  }
  \end{aligned}
\end{equation}
where ${\left( m,C \right)}$ are mean and variance  of the generated data, and ${ \left( m _{r}, C_{r} \right)}$  are mean and variance of the real data.

The R-precision is to evaluate whether the synthesized image is consistent with the corresponding description by retrieving the text description for a given image. For a fair comparison, we follow the evaluation settings with 
DMGAN ~\cite{DMGAN} and quote their results. The VLMS is calculated by a pre-trained VLM model, considering both the image quality and the visual-textual semantic consistency. 

\subsection{Implementations}
The proposed dual vision-language matching strategy on two baselines is implemented with 7700k CPU and 8 NVIDIA GeForce GTX2080ti GPUs. For the vision encoder, text encoder and visual-language matching scoring model, the batch size is set to 64. For the $\text{VLMGAN}_{+\text{AttnGAN}}$ model, the batch size is set to 24, and the learning rate is 0.0001 for the generators and 0.0004 for the discriminators. We only apply the dual multi-level vision-language supervision in the last generator (256x256) due to the low-resolution images are not well synthesized. The ADAM optimizer ~\cite{adam} is adopted to optimize the proposed model. The training of $\text{VLMGAN}_{+\text{AttnGAN}}$ is stopped at 600 epochs for CUB bird dataset and at 120 epochs for MSCOCO dataset following previous works ~\cite{AttnGAN, DMGAN}. The parameters of generative networks and the discriminating networks are optimized alternatively. For ${\text{VLMGAN}_{+\text{AttnGAN}}}$, the training process is shown in Algorithm ~\ref{al1}. For another baseline (DFGAN), the settings are similar with those of ${\text{VLMGAN}_{+\text{AttnGAN}}}$. 

\begin{algorithm}[htb]
  \small
  \caption{Training procedure of $\text{VLMGAN}_{+\text{AttnGAN}}$}
  \label{alg:Framwork}
  \begin{algorithmic}[1]
    \Require
      Pre-trained models (Vision-Encoder, Text-Encoder, and MSB);  Batch size $M$; Text-image paired instances $\{T, I\}$; Learning rate $\alpha$.
    \Ensure
      Generators ($G_0$, $G_1$, $G_2$) and discriminators ($D_0$, $D_1$, $D_2$);
    \Repeat
    \State Sample image-text pairs ${\{I_i, T_i\}}$ and generate random noise vector ${z_i}$;
    \State Extract representation of text captions by
    ${  \varphi, \overline { \varphi } = F_{\text{Text-Encoder}}(w_1, w_2, ..., w_n)}$;

    \State Generate fake images by $ {( \hat{x}_0, \hat{x}_1, \hat{x}_2)}\leftarrow G({\overline { \varphi }}, {\varphi}, {z_i})$ 
        \For{$i \in [0,1,2]$}
           
          \State Calculate the discriminative loss ${\mathcal{L}_{{D_{{i}}}}}$:
          \State Update parameters of discriminator $D_i$ by Adam optimizer; 
        \EndFor
        \For{$i \in [0,1, 2]$}
          \State Calculate the generative loss $\mathcal{L}_{G_i}$;
          \If {$i$ is equal to 2}
          \State Calculate textual-visual matching loss $\mathcal{L}_{VLM}$;
          \State Calculate visual-visual matching loss $\mathcal{L}_{VVM}$;
          \EndIf
        \EndFor
    \State Calculate total loss $\mathcal{L}_{G}$:
    \Statex  \qquad  \qquad $\mathcal{L}_{G} \leftarrow \mathcal{L}_{G_0} + \mathcal{L}_{G_1} + \mathcal{L}_{G_2}+ \mathcal{L}_{VVM}+ \mathcal{L}_{VLM}$;
    \State Update the parameters of the generators ($G_0$, ${G_1}$, $G_2$ ) by Adam optimizer;
  \Until $\text{VLMGAN}_{+\text{AttnGAN}}$ converges\\
  \Return $G_0$, $G_1$, $G_2$, $D_0$, $D_1$, $D_2$.
  \end{algorithmic}
  \label{al1}
\end{algorithm}

\begin{table}[t]\setlength{\tabcolsep}{6.0pt}
   \small
   \caption{The Inception Score comparison of the proposed VLMGAN* and the state-of-the-art methods on the CUB bird dataset and MSCOCO dataset. DFGAN* means the scores are obtained by using their pre-trained model. }\label{table1}
   \begin{tabular}{lllllll}
      \toprule
      Model & Resolution & CUB & MSCOCO\\
      \midrule
      GAN-INT-CLS ~\cite{GAN-INT-CLS} & 64x64 & 2.88 ± 0.04 & 7.88 ± 0.07\\
      GAWWN ~\cite{GAWWN} & 256x256  & 3.62 ± 0.07 & -\\
      StackGAN ~\cite{StackGAN} & 256x256  & 3.70 ± 0.04 & 8.45 ± 0.03\\
      StackGAN++ ~\cite{StackGAN++} & 256x256  & 3.82 ± 0.06 & -\\
      HDGAN ~\cite{HDGAN} & 512x512  & 4.15 ± 0.05 & -\\
      
      MirrorGAN ~\cite{MirrorGAN}& 256x256  & 4.56 ± 0.04 & 26.47 ± 0.41\\
      LeicaGAN ~\cite{LeicaGAN}& 256x256  & 4.62 ± 0.06 & -\\
      DMGAN ~\cite{DMGAN} & 256x256  & 4.75 ± 0.07  & 30.49 ± 0.57\\
      Bridge-GAN ~\cite{Bridge-GAN} & 256x256 & 4.74  ± 0.04 & 16.40  ± 0.30 \\
      OPGAN ~\cite{OP-GAN}  & 256x256  & - & 28.57 ± 0.17 \\
      C4Synth ~\cite{c4synth} & 256x256 & 4.07 ± 0.13 & -\\
      CGL-GAN ~\cite{cglgan}& 256x256 & 3.67 ± 0.04 & 13.62 ± 0.02 \\
      KTGAN ~\cite{ktgan} &
      256x256 & 4.85±0.04 & 31.67 ± 0.36 \\
      LD-CGAN ~\cite{ldgan} &128x128& 4.18± 0.06 & - \\
      SAMGAN ~\cite{SAMGAN} & 256x256 & 4.61 ± 0.03 & 27.31 ± 0.23 \\
      CPGAN$^*$~\cite{cpgan} & 256x256 & - & 52.73 ± 0.61 \\
      MA-GAN ~\cite{MAGAN} & 256x256 & 4.76 ± 0.05 & - \\
      AttnGAN ~\cite{AttnGAN} & 256x256  & 4.36 ± 0.04 & 25.89 ± 0.47\\
      DFGAN ~\cite{DFGAN} & 256x256 & 4.86 ± 0.04 & - \\
      DFGAN* ~\cite{DFGAN} & 256x256 & 4.70 ± 0.05 & 18.70 ± 0.07  \\
      VLMGAN$_{+\text{AttnGAN}}$ & 256x256 & 4.86 ± 0.06 & 31.84 ± 0.46 \\
      VLMGAN$_{+\text{DFGAN}}$ & 256x256 & 4.95 ± 0.04 & 26.51 ± 0.43 \\
      \bottomrule
   \end{tabular}
\end{table}

\begin{table}[t]\setlength{\tabcolsep}{5.5pt}
    \centering
    \caption{The FID and R-precision comparison of AttnGAN, DMGAN, DFGAN, $\text{VLMGAN}_{+\text{AttnGAN}}$ and $\text{VLMGAN}_{+\text{DFGAN}}$ on CUB dataset and MSCOCO dataset. DFGAN* means the scores are obtained by using their released pre-trained model. The `$\downarrow$' means the lower, the better.  The `$\uparrow$' means the higher, the better. }
    \begin{tabular}{l|cc|cc}
        \toprule
        \multirow{2}*{\bfseries Methods}  &
        \multicolumn{2}{c}{\bfseries CUB} & \multicolumn{2}{c}{\bfseries MSCOCO}\\

        \cmidrule{2-5} &FID $\downarrow$ &R-precision $\uparrow$ &FID $\downarrow$ &R-precision $\uparrow$\\
        \midrule
        AttnGAN \cite{AttnGAN}&23.98&67.82±4.43&35.49&85.47±3.69\\
        DMGAN \cite{DMGAN}&16.09&72.31±0.91&32.64&88.56±0.28\\
        DFGAN \cite{DFGAN}&19.24&-&28.92&-\\
        DFGAN* \cite{DFGAN}&21.85  & 38.76±0.08&27.39 &55.34±0.90\\
        VLMGAN$_{+\text{AttnGAN}}$&15.02&77.75±0.74&31.24&89.45±0.52\\
        VLMGAN$_{+\text{DFGAN}}$&16.04&72.59±0.32&23.62&82.95±0.60\\
        \midrule
    \end{tabular}
    \centering
    \label{fid_r}
\end{table}

\subsection{Quantitative Evaluation}
The proposed dual vision-language matching module can be applied to other text-to-image synthesis architectures. In our experiments, we apply the dual vision-language matching module on two popular baselines, AttnGAN and DFGAN. They are marked as ${\text{VLMGAN}_\text{+AttnGAN}}$ and ${\text{VLMGAN}_\text{+DFGAN}}$ respectively. 

The experimental results of Inception Score on CUB and MSCOCO are reported in Table ~\ref{table1}. 
Table ~\ref{table1} shows that the proposed ${\text{VLMGAN}_\text{+AttnGAN}}$ achieves 4.86 on the CUB bird dataset, which outperforms the other methods except for DFGAN and ${\text{VLMGAN}_\text{+DFGAN}}$. 
Compared with baseline (AttnGAN) ~\cite{AttnGAN}, the proposed method ${\text{VLMGAN}_\text{+AttnGAN}}$ can improve the Inception Score from 4.36 to 4.86 and 25.89 to 31.84. This suggests that the ${\text{VLMGAN}_\text{+AttnGAN}}$ can generate images with better diversity and image quality. It should be noted that the IS value of the proposed ${\text{VLMGAN}_\text{+AttnGAN}}$ is lower than CPGAN ~\cite{cpgan}, as a result of CPGAN adopts pre-trained Yolo-v3 ~\cite{yolov3} as the discriminator.  With extra information, CPGAN obtains the highest IS score on MSCOCO. The proposed ${\text{VLMGAN}_\text{+AttnGAN}}$ can obtain same results with DFGAN on CUB bird daraset. In addition to AttnGAN, we also choose DFGAN as baseline to implement the dual vision-language matching strategy named ${\text{VLMGAN}_\text{+DFGAN}}$. In original paper of DFGAN, the authors do not report the IS and R-precision on MSCOCO dataset. Therefore, we obtain the values by using their public available pre-trained model, which is marked with DFGAN*. The results in Table ~\ref{table1} indicate that ${\text{VLMGAN}_\text{+DFGAN}}$ obtain the best IS score (4.95). However, the IS score on MSCOCO dataset is lower than other methods. This phenomenon is also appeared in Bridge-GAN \cite{Bridge-GAN}. The cause may be DFGAN only adopts the sentence feature as condition and  the diversity of synthesized images decreases without word features. Compared with its baseline (DFGAN), ${\text{VLMGAN}_\text{+DFGAN}}$ can improve the IS from 18.70 to 26.51. In summary, the proposed vision-language matching strategy is beneficial to the performance of two baselines.      


The FID and R-precision of AttnGAN, DMGAN, DFGAN, ${\text{VLMGAN}_\text{+AttnGAN}}$, and ${\text{VLMGAN}_\text{+DFGAN}}$ on CUB dataset and MSCOCO dataset are reported in Table ~\ref{fid_r}. DMGAN, DFGAN, and VLMGAN* are also improved from AttnGAN. Compared with AttnGAN, the proposed ${\text{VLMGAN}_\text{+AttnGAN}}$ outperforms it by a large margin on both two datasets. Specifically, ${\text{VLMGAN}_\text{+AttnGAN}}$ improves R-precision from 67.82 to 77.75 for CUB and from 85.47 to 89.45 for MSCOCO. R-precision evaluates whether the synthesized image is consistent with the text description by retrieving manner. The image feature for retrieval is obtained by the pre-trained Vision-Encoder. The results show that the proposed dual vision-language matching strategy makes a pivotal contribution to improve visual-textual semantic consistency. The comparison of the FID score indicates that the image distribution generated by ${\text{VLMGAN}_\text{+AttnGAN}}$ is closer to real image distribution than others on CUB dataset. Comparing to advanced DMGAN and DFGAN, the proposed ${\text{VLMGAN}_\text{+AttnGAN}}$ also keeps its superiority,which can obtain competitive results. For another baseline (DFGAN), we can find that ${\text{VLMGAN}_\text{+DFGAN}}$ obtains the relatively lower FID on CUB dataset and the lowest FID on MSCOCO dataset. The FID scores of DFGAN shows that DFGAN have excellent performance in synthesizing photo-realistic image. However its semantic consistency is relatively poor in term of R-precision. This phenomenon also verifies that some methods can not be good at both image reality and text semantic consistency. By strengthening the vision-language matching, ${\text{VLMGAN}_\text{+DFGAN}}$ can improve the performance on both FID and R-precision.

\subsection{Qualitative Evaluation}

\begin{figure*}
   \begin{center}
   \centering
   \includegraphics[width=1\linewidth]{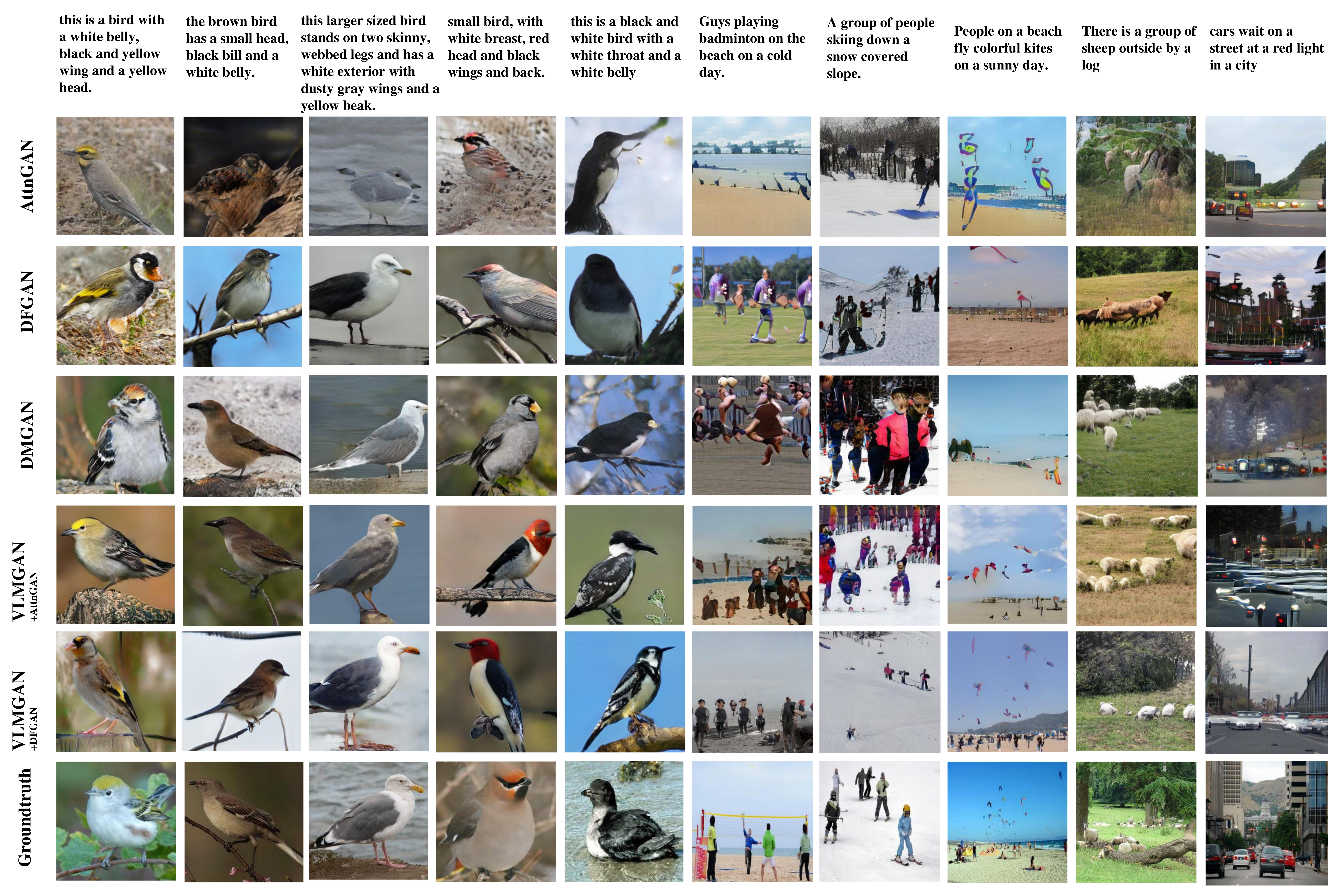}
   \caption{Examples synthesized by AttnGAN ~\cite{AttnGAN}, DFGAN ~\cite{DFGAN},
    DMGAN ~\cite{DMGAN}, ${\text{VLMGAN}_{+\text{AttnGAN}}}$, and ${\text{VLMGAN}_{+\text{DFGAN}}}$.  The images in the last row are the corresponding ground truth. The images in the same column are conditioned the same description.} 
  \label{figure4}
  \end{center}
\end{figure*}

%

Visual comparisons of AttnGAN, DMGAN, DFGAN, ${\text{VLMGAN}_{+\text{AttnGAN}}}$, and ${\text{VLMGAN}_{+\text{DFGAN}}}$ are shown in Figure ~\ref{figure4}. In this paper, we use `VLMGAN*' to denote the dual-matching driven methods ${\text{VLMGAN}_{+\text{AttnGAN}}}$ and ${\text{VLMGAN}_{+\text{DFGAN}}}$. In general, the images synthesized by ${\text{VLMGAN}_{+\text{AttnGAN}}}$ and ${\text{VLMGAN}_{+\text{DFGAN}}}$ are more realistic and highly consistent with the text description because it employs a vision-language matching model and a visual consistency constraint. On the CUB dataset, the proposed model can better understand the description and synthesize a more clearly structured image. Comparing with MSCOCO, the CUB is more straightforward, so that all of these methods have relatively better performance. In terms of complex image synthesis, the MSCOCO dataset is adopted to verify the proposed method's performance. The models with vision-language matching  strategy can  precisely understand the text description and generates a well-structured image. For example, VLMGAN* well presents the shape and structure of kites like the ground-truth, while other methods can not.   The visual comparison shows that the proposed VLMGAN* has superiority in keeping semantic and visual consistency by using a multi-level vision-language matching model and a visual-consistent constraint. 

\begin{table}[t]\setlength{\tabcolsep}{4.6pt}
   \caption{The performance of different components of the proposed VLMGAN on CUB dataset.`w/o' means 'without'. }\label{table3}
   \begin{tabular}{lllllll}
      \toprule
      Architectures & FID  $\downarrow$& IS  $\uparrow$& R-precision $\uparrow$\\
      \midrule
      AttnGAN, w/o DAMSM & 53.73 & 3.89 ± 0.04 & 10.37 ± 5.88 \\
      AttnGAN (baseline) & 23.89 & 4.36 ± 0.03 & 67.82 ± 4.43\\
      ${\text{VLMGAN}_{+\text{AttnGAN}}}$, w/o VLM & 33.25   & 4.20 ± 0.05 & 46.45 ± 3.36\\
      ${\text{VLMGAN}_{+\text{AttnGAN}}}$, w/o VVM & 16.23   & 4.73 ± 0.07 & 73.56 ± 0.82\\
      ${\text{VLMGAN}_{+\text{AttnGAN}}}$ & 15.02  & 4.86 ± 0.06 & 77.75 ± 0.74 \\
      \bottomrule
   \end{tabular}
\end{table}

\begin{table}[t]\setlength{\tabcolsep}{4.6pt}
\caption{Ablation studies on different level matching on CUB bird dataset. `w/o' means 'without'.}
    \centering
    \begin{tabular}{l|ccc}
        \toprule
        {\bfseries Model Setting}&FID $\downarrow$ &IS $\uparrow$&R-precision $\uparrow$\\
        \midrule
        Baseline &53.73&3.89 ± 0.04&10.37 ± 5.88\\
        ${\text{VLMGAN}_{+\text{AttnGAN}}}$, w/o General&17.00&4.76 ± 0.07&73.41 ± 0.60\\
        ${\text{VLMGAN}_{+\text{AttnGAN}}}$, w/o Local&23.32&4.47 ± 0.07&57.40 ± 0.88\\
        ${\text{VLMGAN}_{+\text{AttnGAN}}}$, w/o Global&18.75&4.65 ± 0.06&67.76 ± 0.56\\
        ${\text{VLMGAN}_{+\text{AttnGAN}}}$&15.20&4.86 ± 0.06&77.75 ± 0.74\\
        \midrule

    \end{tabular}
    \label{match-ablation}
\end{table}

\subsection{Ablation Study}

To thoroughly verify VLMGAN*'s effectiveness, we do ablation studies on VLMGAN* and its variants. VLMGAN* means the dual vision-language matching strategy based GAN for text-to-image synthesis.
We conduct this ablation study of on ${\text{VLMGAN}_{+\text{AttnGAN}}}$ on CUB dataset.
Several comparative experimental results are reported in Table ~\ref{table3}. ``VLM''  means the proposed vision-language matching constraint, and ``VVM'' means the visual-visual matching consistent constraint between the synthesized image and original real image. The AttnGAN is our baseline. The comparing results show that the vision-language matching model is fundamental in improving the image quality. Without the VLM model, the performance of all evaluation metrics drop rapidly , such as the IS score drops from 4.86 to 4.20, the FID increases from 15.02 to 33.25, the R-precison drops from 77.75 to 46.45. Comparing ``${\text{VLMGAN}_{+\text{AttnGAN}}}$, w/o VVM'' to ``AttnGAN'', the results indicate that the proposed multi-level vision-language model, which can effectively match the image content and the textual semantic information, is better than DAMSM. Besides, the visual consistency constraint between the synthesized and the real image also can improve the scores by a considerable margin. These experimental results show that the components of the proposed method contribute to improving the image quality.

To clarify the contribution of different level matching, we conduct more experiments on CUB dataset. We also adopt the AttnGAN as the baseline. The experimental results are shown in Table \ref{match-ablation}. We can find that the local-level matching makes the biggest contribution to the performance, which improves the IS from 4.47 to 4.86, FID from 23.32 to 15.20. The three kinds of matching can improve the model performance in terms of different metrics. 

The above ablation studies can verify the effectiveness of the proposed dual multi-level vision-language matching strategy.

\subsection{Effectiveness of VLMS}

\begin{figure}
   \begin{center}
   \centering
   \includegraphics[width=1\linewidth]{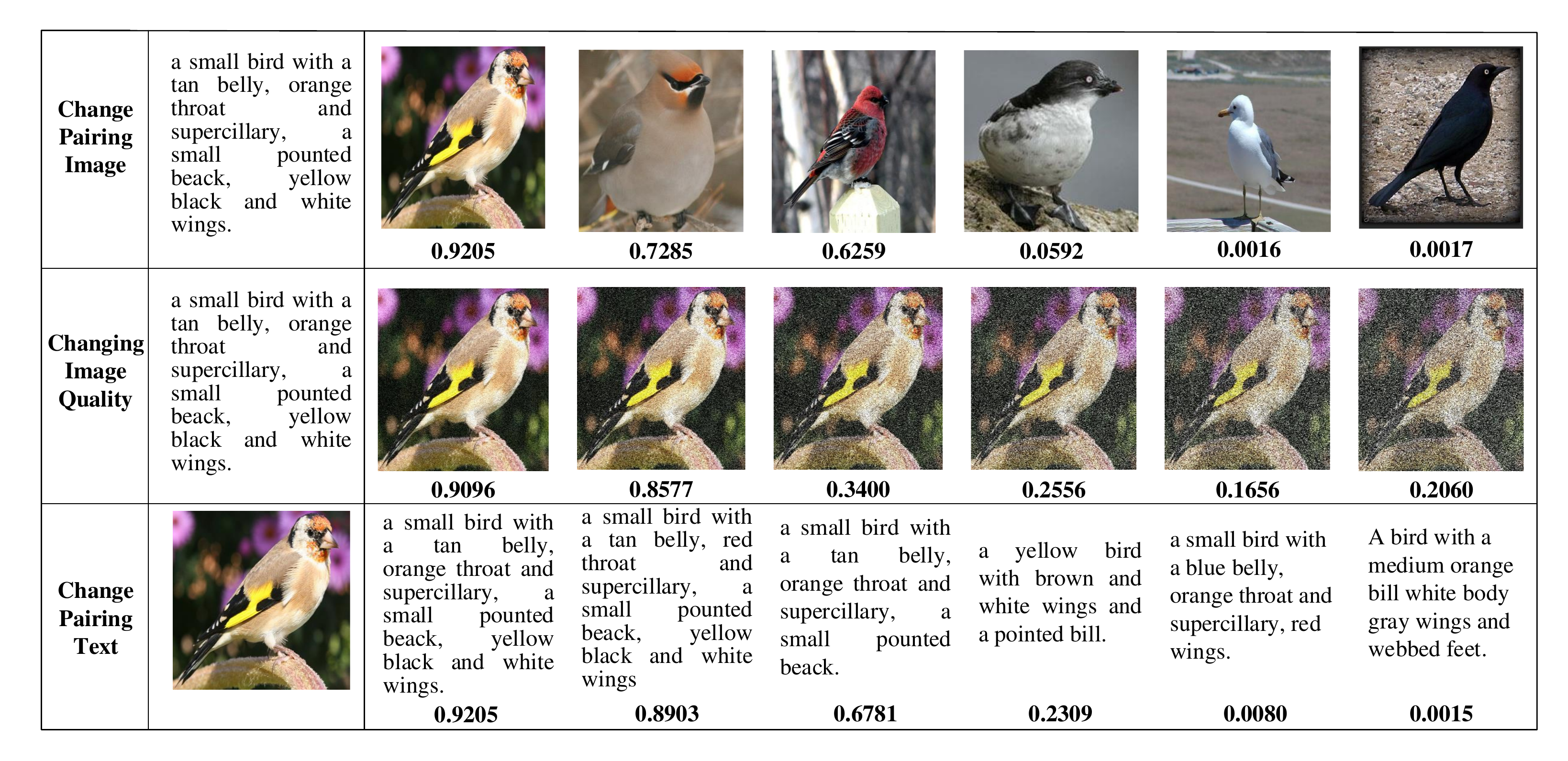}
   \caption{Some examples of VLMS by modifying the image and text description. The numbers under the images ( texts) are the corresponding VLMS value between the left  text (image).}
  \label{figure_vlms}
  \end{center}
\end{figure}

\begin{table}[t]\setlength{\tabcolsep}{20.0pt}
   \centering
   \caption{The VLMS variation by changing the pairing image quality on CUB dataset.}\label{table_var}
   \begin{tabular}{cc}
      \toprule
       Settings& VLMS $\uparrow$\\
      \midrule
      Ground truth&0.77±0.30\\
      Random image& 0.12±0.22\\
      $\sigma(0.01)$&0.63±0.29\\
      $\sigma(0.1)$&0.40±0.30\\
      $\sigma(0.3)$&0.23±0.23\\
      $\sigma(0.5)$&0.21±0.21\\
      $\sigma(1)$&0.18±0.20\\
      \bottomrule
      \label{noise_verify}
   \end{tabular}
\end{table}

\begin{table}[t]\setlength{\tabcolsep}{20.0pt}
   \centering
   \caption{The VLMS variation by changing the pairing text description on CUB dataset.}\label{table_var}
   \begin{tabular}{cc}
      \toprule
       Settings& VLMS  $\uparrow$\\
      \midrule
      Ground truth&0.77±0.30\\
      Random text& 0.12±0.22\\
      mask stopwords & 0.75±0.24\\
      10\%&0.67±0.26\\
      20\%&0.56±0.27\\
      50\%&0.37±0.26\\
      70\%&0.35±0.25\\
      90\%&0.10±0.22\\
      \bottomrule
      \label{mask_verify}
   \end{tabular}
\end{table}

\begin{table}[t]\setlength{\tabcolsep}{12.0pt}
   \centering
   \caption{The VLMS comparison with different models on CUB dataset.}\label{table_var}
   \begin{tabular}{ccc}
      \toprule
       Methods& CUB&MSCOCO\\
      \midrule
      AttnGAN&0.49±0.20&0.36±0.28\\
      DMGAN& 0.52±0.22&0.42±0.30\\
      DFGAN&0.53±0.27&0.44±0.23\\
      ${\text{VLMGAN}_\text{+AttnGAN}}$&0.55±0.27&0.47±0.26\\
      ${\text{VLMGAN}_\text{+DFGAN}}$&0.56±0.23&0.46±0.23\\
      \bottomrule
      \label{vlms_compare}
   \end{tabular}
\end{table}

In this subsection, we firstly explain the effectiveness and rationality of the proposed novel text-to-image evaluation metric, named Vision-Language Matching Score (VLMS), which directly measures the similarity between the synthesized image and the corresponding text description by the pre-trained Matching Scoring Block of the VLM model. To verify the effectiveness of the proposed VLMS, we conduct experiments with different variants, as shown in Table ~\ref{noise_verify} and Table \ref{mask_verify}. In Table \ref{noise_verify}, `Random text' means that we randomly select a text from the dataset for a specific image, which obtains the lowest VLMS. This result indicates that VLMS is sensitive to the image-text semantic consistency. For changing the image quality, we add different level white gaussian noise (standard deviation: 0.01, 0.1, 0.3, 0.5, 1). We can find that the VLMS decreases rapidly with the increase of noise. This phenomenon indicates that the VLMS is sensitive to the image quality.
For changing the text description, we randomly replace or remove words of the whole sentence by different percentage (10\%, 30\%, 50\%, 70\%, 90\%). For a specific text description, the ground truth obtains the best VLMS score, and the dissimilar images receive relatively low scores. 
Besides, we add experiments to analyze the influence of these irrelevant words, such as `the', `a' and so on. To be specific, we build a stop words dictionary, which contains ‘and, this, a, an, there, of’. If the words of the sentence are in the stop words, we mask them in calculating the VLMS. We find that these irrelevant words make little impact on the final results (from 0.77 to 0.75).
The comparisons show that the lower the image quality, the lower the score. If we modify the text description, the VLMS scores also decrease. For better understanding, we present some examples to explain this, as shown in Figure ~\ref{figure_vlms}. Therefore, from above analyses, the proposed VLMS metric is reasonable to measure the image-text matching score, which considers both image quality and semantic consistency. We calculate the mean VLMS score of 30000 generated images. Table ~\ref{vlms_compare} shows that ${\text{VLMGAN}_\text{+DFGAN}}$ obtains the highest value on CUB datasets and ${\text{VLMGAN}_\text{+AttnGAN}}$  obtains the highest value on MSCOCO datasets.

\subsection{Convergence Analysis}

\begin{figure}
   \begin{center}
   \centering
      \subfigure[CUB]{
      \begin{minipage}[t]{0.5\linewidth}
      \centering
      \includegraphics[height=1.5in,width=1.7in]{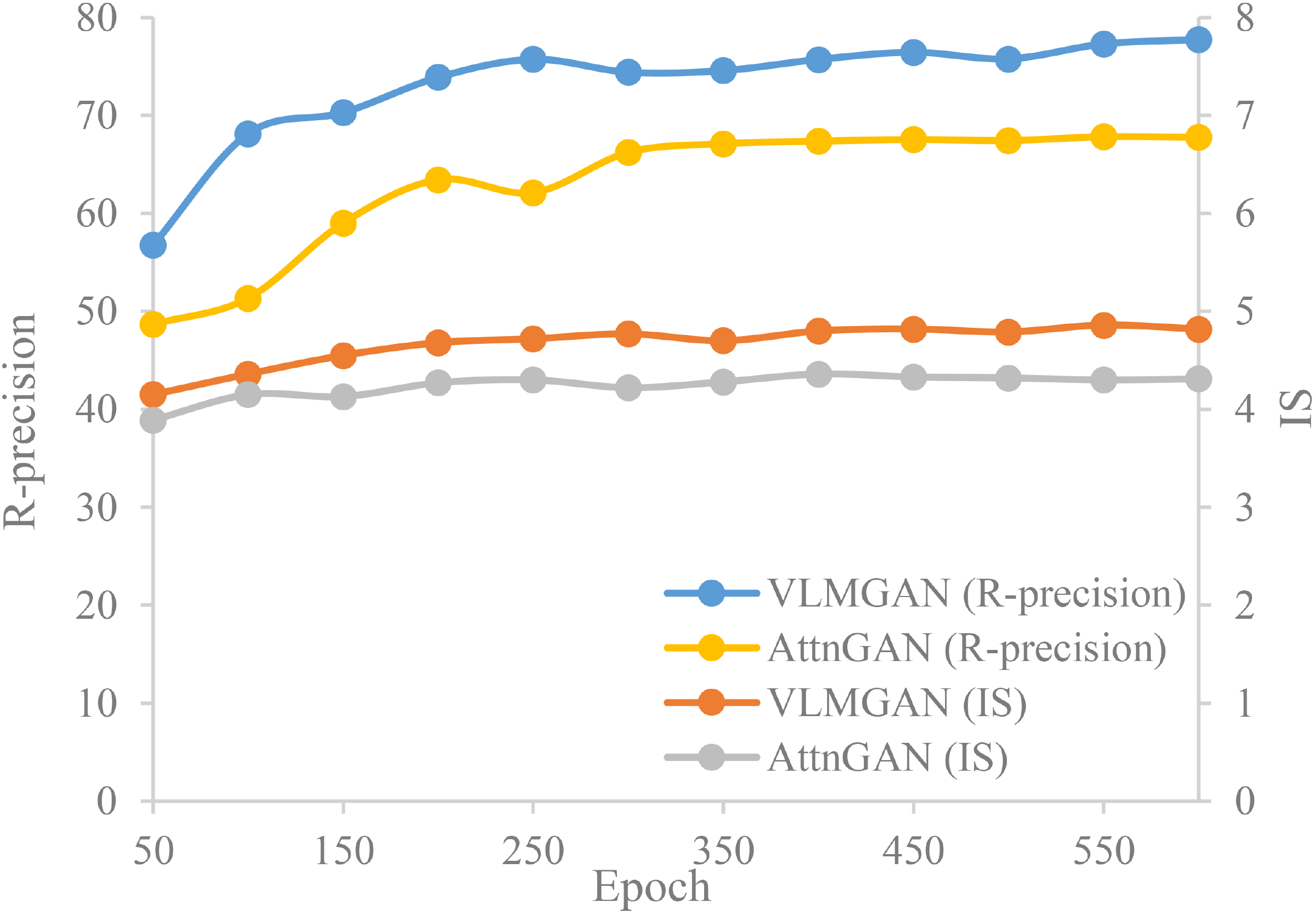}
      \end{minipage}%
      }%
      \subfigure[MSCOCO]{
      \begin{minipage}[t]{0.5\linewidth}
      \centering
      \includegraphics[height=1.5in,width=1.7in]{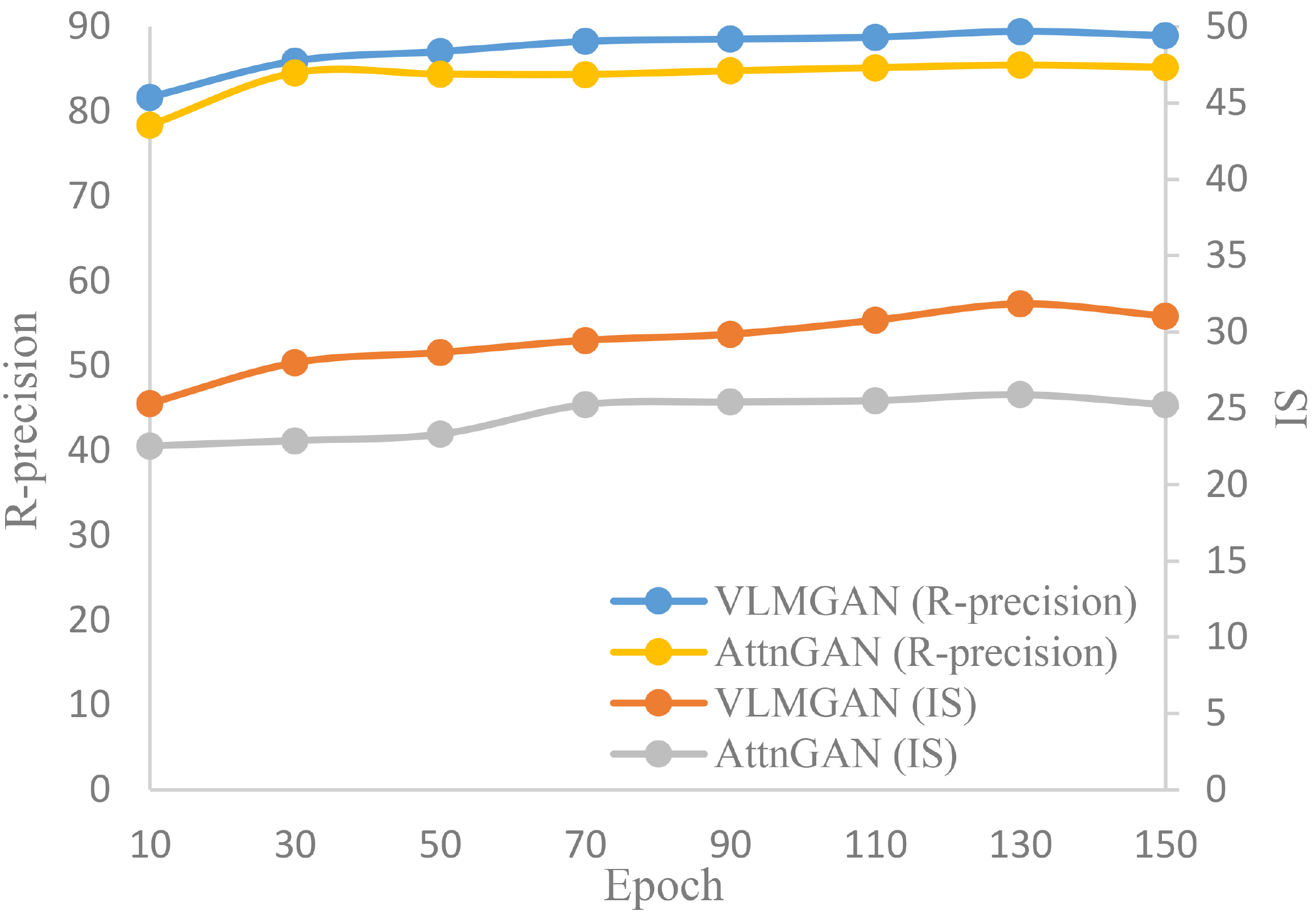}
      \end{minipage}
      }
      \caption{IS and R-precision comparison between AttnGAN ~\cite{AttnGAN} and ${\text{VLMGAN}_{+\text{AttnGAN}}}$ with the training processing.}
      \label{AttnGAN_VLM}
  \end{center}
\end{figure}

\begin{figure}
   \begin{center}
   \centering
      \subfigure[Loss of Generator and Discriminator]{
      \begin{minipage}[t]{0.5\linewidth}
      \centering
      \includegraphics[height=1.6in,width=1.8in]{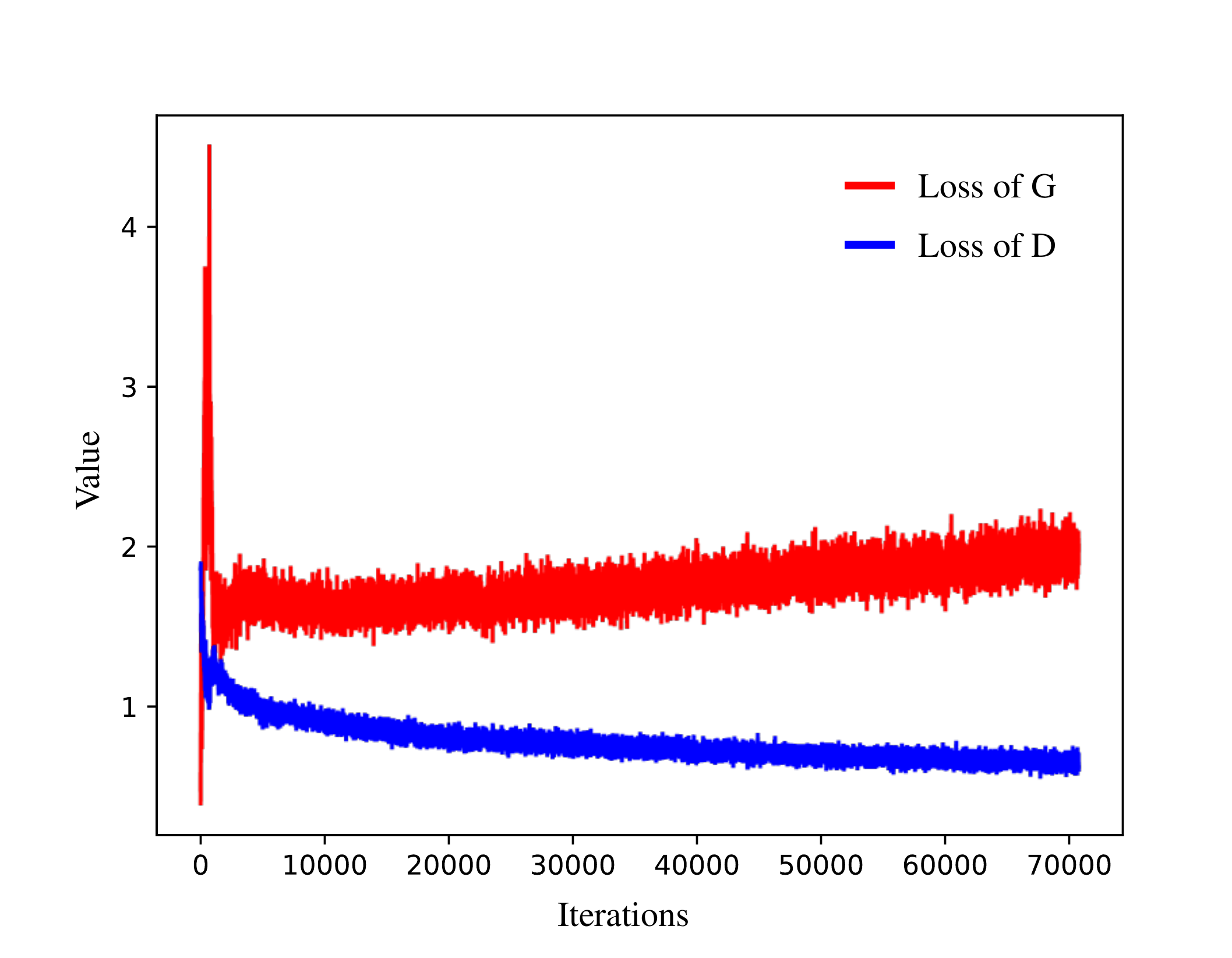}
      \end{minipage}%
      }%
      \subfigure[Loss of VVM and VLM]{
      \begin{minipage}[t]{0.5\linewidth}
      \centering
      \includegraphics[height=1.6in,width=1.8in]{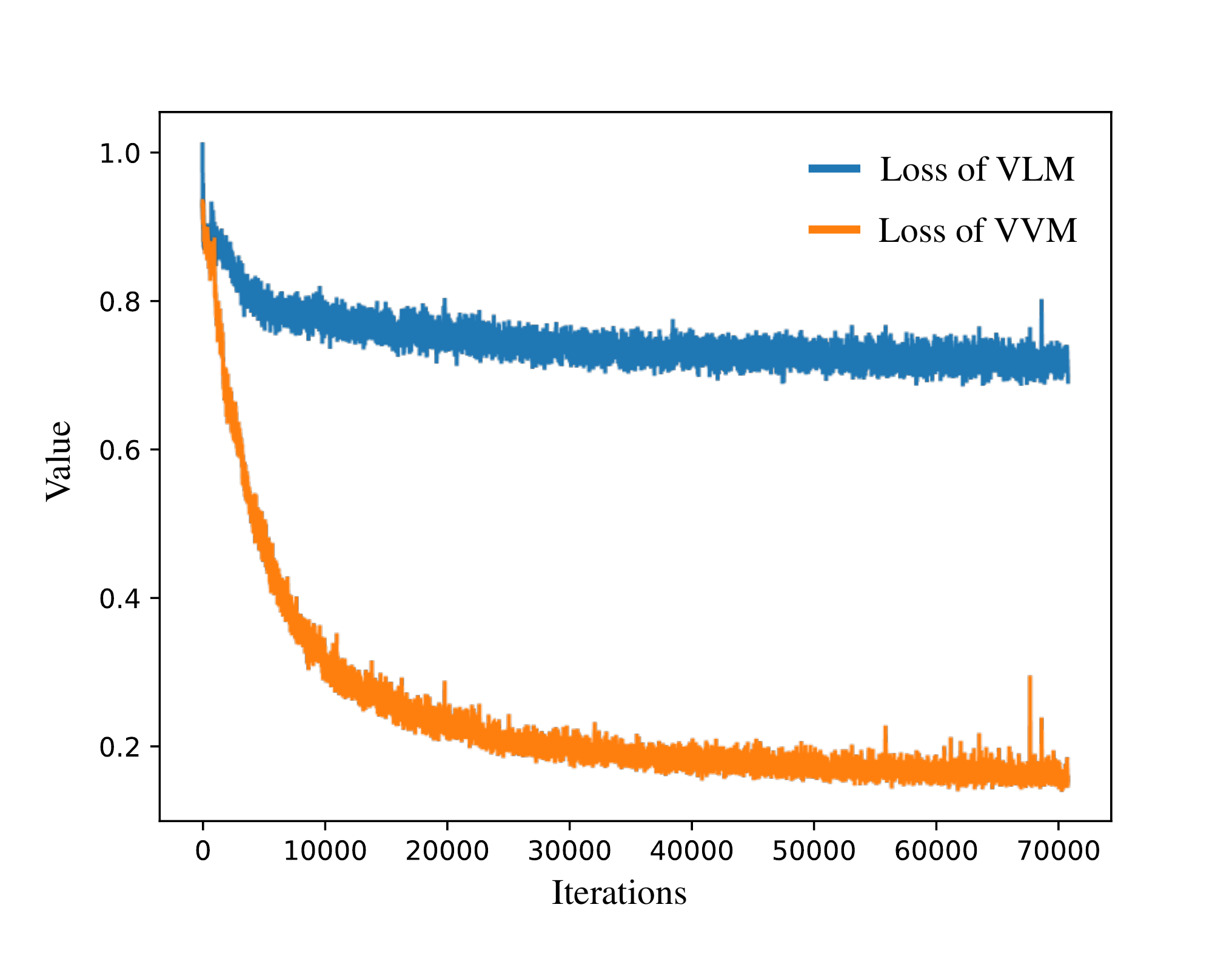}
      \end{minipage}
      }
      \caption{The training loss of ${\text{VLMGAN}_{+\text{DFGAN}}}$.}
      \label{DFGAN_loss}
  \end{center}
\end{figure}

Figure ~\ref{AttnGAN_VLM} shows the comparison between ${\text{VLMGAN}_{+\text{AttnGAN}}}$ and AttnGAN with the training processing. This figure shows that ${\text{VLMGAN}_{+\text{AttnGAN}}}$ exceeds AttnGAN in the whole training process. Besides, the training losses of ${\text{VLMGAN}_{+\text{DFGAN}}}$ are shown in Figure \ref{DFGAN_loss}. It should be noted that loss of generator does not include the loss of $\mathcal{L}_{VVM}$ and $\mathcal{L}_{VLM}$. By comparing $\mathcal{L}_{VVM}$ and $\mathcal{L}_{VLM}$, we can find that the decrease of $\mathcal{L}_{VVM}$ is more significant. The model is converged after about 70000 iterations.

\subsection{Generalization Study}
\begin{figure}
   \begin{center}
   \centering
   \includegraphics[width=1\linewidth]{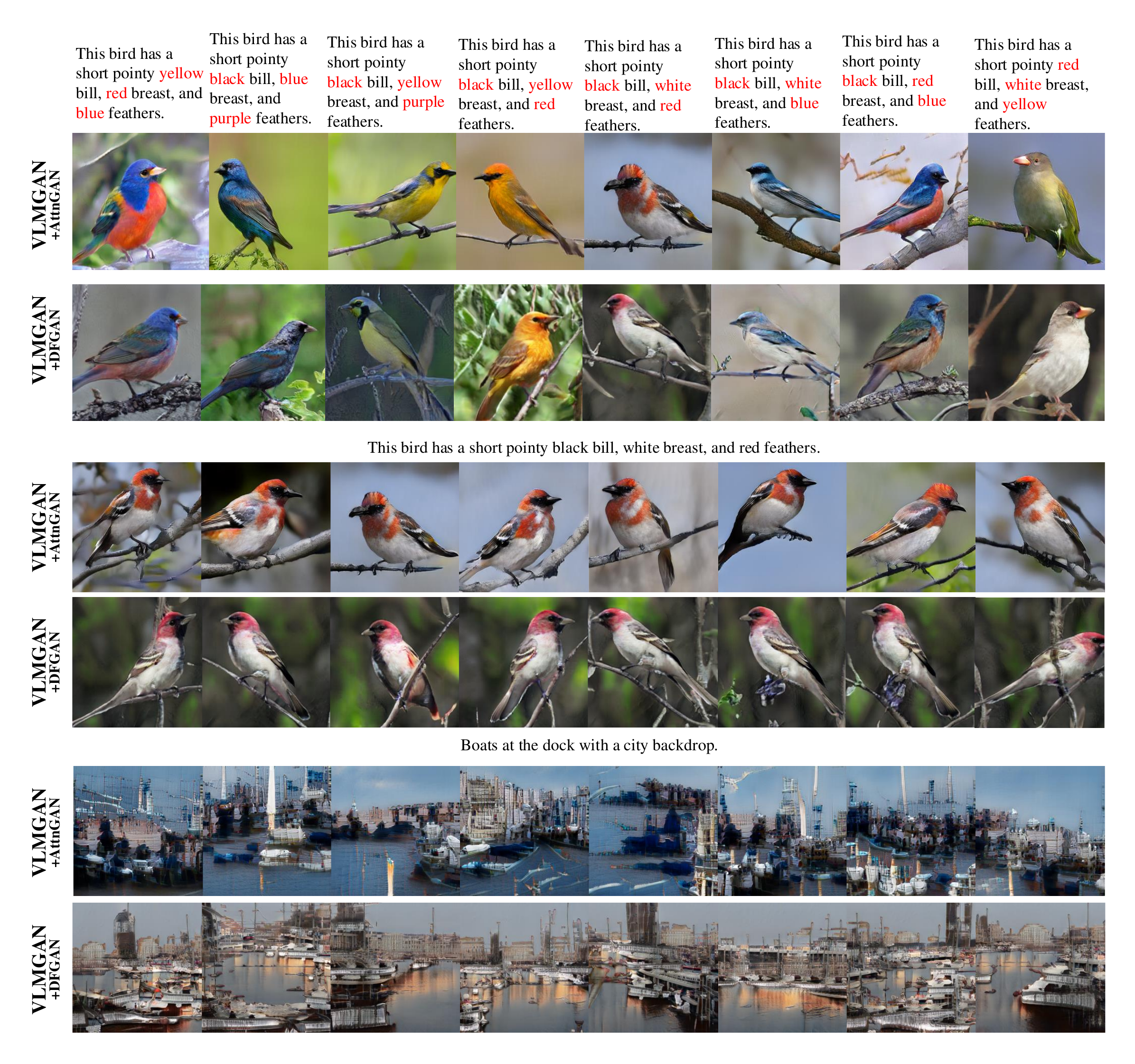}
   \caption{More synthesized examples of ${\text{VLMGAN}_{+\text{AttnGAN}}}$ and ${\text{VLMGAN}_{+\text{DFGAN}}}$.}
  \label{figure5}
  \end{center}
\end{figure}

In this section, we conduct more experiments to analyze the generalization and robustness of the proposed approach. The first experiment is modifying some attribute words when generate the corresponding image, as shown in the top row of Figure ~\ref{figure5}. The words in red are some key attributes when describing the bird. We randomly replace these words with other attributes words. The results of modifying some important words show that the synthesized images can keep consistency with the descriptions' variation. The second experiment is generating a series of images by fixing the text description.  The results of some examples conditioned the same text of CUB dataset and MSCOCO dataset are presented in the second and three rows of Figure ~\ref{figure5}, respectively. From the second row, we can find that the attributes of the synthesized birds (black bill, white breast and red feathers) keep consistency among different images. In the same time, we can observe that these birds have rich variety of the posts, such as the direction of body. This experimental phenomenon is also appeared on the MSCOCO dataset. Therefore, we can conclude that the proposed method VLMGAN* have excellent generalization and robustness.      

\section{Conclusion}
\label{conclusion}
This paper addresses the text-to-image synthesis by strengthening the semantic and visual matching between the synthesized image and the real data. 
To this end, the proposed dual multi-level vision-language matching considers both textual-visual matching and visual-visual matching. By introducing this idea into generative architecture, the VLMGAN* successfully exploits this idea and achieves excellent image quality performance. In addition, the VLM can also measure the matching score between the image and text by considering both image quality and image semantic, which is more consistent with our human perception. We implement the proposed dual vision-language matching strategy on two popular baselines, AttnGAN and DFGAN. The experimental results of ${\text{VLMGAN}_{+\text{AttnGAN}}}$ and ${\text{VLMGAN}_{+\text{DFGAN}}}$ show that the VLMGAN* achieves state-of-the-art performance on CUB dataset and more challenging MSCOCO dataset. Compare with the baselines, both ${\text{VLMGAN}_{+\text{AttnGAN}}}$ and ${\text{VLMGAN}_{+\text{AttnGAN}}}$ can significantly improve their performance. In the future study, we will try to explore more excellent mechanisms to improve the quality of synthesized image and semantic consistency between the synthesized image and text.





\ifCLASSOPTIONcaptionsoff
  \newpage
\fi



\bibliographystyle{IEEEtran}
\bibliography{ref}
\end{document}